\documentclass[10pt,twocolumn,letterpaper]{article}

\usepackage[compact]{titlesec}

\usepackage{wacv}              %

\usepackage{graphicx}
\usepackage{amsmath}
\usepackage{amssymb}
\usepackage{booktabs}

\usepackage{algorithm}
\usepackage{algpseudocode}
\usepackage{float}
\usepackage{caption}
\usepackage{array}
\usepackage{xcolor}
\usepackage{tablefootnote}
\usepackage{arydshln}
\usepackage{microtype}
\usepackage[T1]{fontenc}
\usepackage{chngcntr}
\usepackage[utf8]{inputenc}

\newcolumntype{C}[1]{>{\centering\arraybackslash}p{#1}}

\usepackage[pagebackref,breaklinks,colorlinks]{hyperref}

\def\wacvPaperID{1179} %
\def\confName{WACV}
\def\confYear{2024}

\begin{document}

\title{Unified Concept Editing in Diffusion Models}
\author{Rohit Gandikota$^{1}$ \qquad Hadas Orgad$^{2}$ \quad Yonatan Belinkov$^{2}$\qquad Joanna Materzy\'nska$^{3}$ \quad David Bau$^{1}$ \vspace{3pt} \\ 
$^{1}$Northeastern University \qquad $^{2}$Technion \qquad $^{3}$Massachusetts Institute of Technology}

\twocolumn[{%
\renewcommand\twocolumn[1][]{#1}%
\maketitle%
\vspace{-0.3in}%
\begin{center}
    \centering\includegraphics[width=\linewidth]{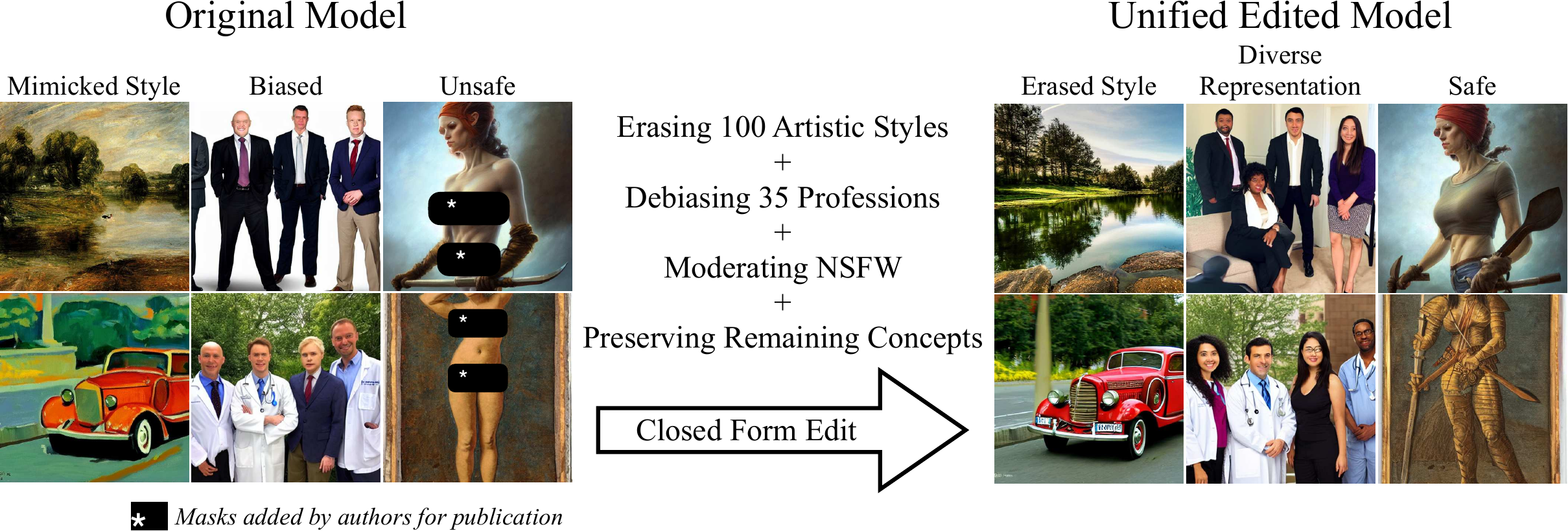}
    \vspace{-0.05in}
    \captionof{figure}{
    Our method enables unified and efficient editing of multiple concepts in text-to-image models through closed-form modifications to attention weights. We present applications to debias, erase, and moderate concepts at scale. Debiasing professions leads the edited model to generate fairer gender and race ratios. Erasing an artistic style removes characteristics associated with a particular creator. Moderating the model reduces the likelihood of generating inappropriate images.
    }
    \label{fig:mainfig}
\end{center}
}]

\maketitle

\begin{abstract}
Text-to-image models suffer from various safety issues that may limit their suitability for deployment.
Previous methods have separately addressed individual issues of bias, copyright, and offensive content in text-to-image models. However, in the real world, all of these issues appear simultaneously in the same model.
We present a method that tackles all issues with a single approach. Our method, Unified Concept Editing (UCE), edits the model without training using a closed-form solution, and scales seamlessly to concurrent edits on text-conditional diffusion models.

We present scalable simultaneous debiasing, style erasure, and content moderation by editing text-to-image projections, 
and perform extensive experiments demonstrating improved efficacy and scalability over prior work.  
Our code is available at \href{https://unified.baulab.info/}{unified.baulab.info}.
\end{abstract}
\vspace{-1em}
\section{Introduction}
    \label{sec:intro}

Text-to-image diffusion models have ushered in a set of complex societal challenges. Generative image models jeopardize artists by cloning their styles\cite{andersen2023stability}; they reinforce biases by amplifying stereotypes\cite{luccioni2023stable, struppek2022biased}; and they facilitate the creation of offensive images\cite{tatum2023porn}. While several methods have been proposed to mitigate such issues individually \cite{chuang2023debiasing, gandikota2023erasing, kumari2023ablating, schramowski2023safe, wang2023concept}, real-world deployments of generative image models manifest all these problems concurrently.
A natural first step for exercising safety in generative models is the careful curation of training data to exclude any content that should not be replicated\cite{rombach2022sd20}.
However, training a large model is expensive, and the impact of data curation on a model may be counterintuitive and unpredictable.
For example, removing undesired content can expose other undesired content~\cite{carlini2022privacy}; removing toxic content can introduce new biases~\cite{dixon2018measuring}; and reducing offensive content can result in incomplete removal~\cite{oconnor2022stable}; 
these examples highlight the limitations of relying solely on data curation.

In this paper, we introduce a unified model-editing approach capable of addressing the different safety issues with a single formulation. Our method, called \emph{Unified Concept Editing} (UCE), offers a fast and practical way to control model behavior post-training, filling the gaps where data curation might fall short. UCE is a closed-form parameter-editing method that enables the application of hundreds of editorial modifications within a single text-to-image synthesis model while preserving the generative quality of the model for unedited concepts.

The UCE method builds upon previous model editing work, generalizing the TIME \cite{orgad2023editing} and MEMIT \cite{meng2022mass} methods. Unlike previous diffusion model editing methods such as TIME, UCE is designed to enable many simultaneous edits to be applied at once. 
These edits can include actions such as erasing, moderating, or debiasing a concept—tasks that have been traditionally treated as distinct issues with separate solutions. 
UCE goes beyond MEMIT in several ways: it edits text-to-image models rather than language models; and it also allows the editor to explicitly specify the distribution of concepts that should not be modified.  Finally, UCE also introduces a new, scalable debiasing approach. 
We compare UCE with a range of prior model-editing methods and find that it demonstrates superior performance, outperforming other methods by a wide margin. UCE exhibits superior performance both in single edits in each category of editing, as well as in the ability to scale to many edits at once while minimizing interference with unedited concepts.

\section{Related Work}
    
While text-to-image diffusion models are becoming increasingly popular in commercial art and graphic design, they tend to suffer from various issues, which have previously been addressed separately.

\paragraph{Copyright issues.} Recent lawsuits \cite{andersen2023stability, setty2023suit} have contended that models like Stable Diffusion infringe on many artistic styles, and researchers have found that the models can memorize some copyrighted training data nearly verbatim~\cite{somepalli2023diffusion, carlini2023extracting}. To reduce such memorization, previous work proposes randomizing and augmenting training image captions~\cite{somepalli2023understanding}, while other work has explored a technique called image cloaking that allows artists to protect their content from being imitated by large generative models by adding specially crafted adversarial perturbations to images before publishing them online~\cite{salman2023raising,shan2023glaze}; both these approaches require thorough preparation of the training corpus.  Another approach adjusts a model after training is complete, deleting an undesired concept by modifying model weights \cite{gandikota2023erasing, kumari2023ablating, kim2023towards,heng2023selective,zhang2023forget}. Our method adopts that concept-erasure approach, and we benchmark against the previous state-of-the-art. Our method differs from previous concept erasure methods because it is a closed-form edit that removes many concepts at once.

\paragraph{Offensive content.} Diffusion models also sometimes generate inappropriate images, such as nude and violent images. Various methods have been proposed to filter out inappropriate images from the training data or at inference time \cite{gandhi2020scalable, nichol2021glide}; for example the Stable Diffusion implementation includes a ``not safe for work'' safety checker that returns a black image when an unsafe image is detected\cite{bedapudi2022nudenet, laborde2022nsfw, rando2022red}, and other work has addressed the issue in through image editing at inference time \cite{schramowski2023safe}. In cases where open-source code and model weights are openly available, such post-production filters can be easily disabled\cite{smith2022howto}. A more difficult-to-circumvent approach removes the knowledge of unwanted concepts from the model weights; previous methods taking that approach have proposed attention re-steering through fine-tuning \cite{zhang2023forget}, fine-tuning the attention weights \cite{gandikota2023erasing} and continual learning \cite{heng2023selective}. While previous methods all fine-tune the model, we propose a fast and efficient method to erase offensive concepts using a closed-form edit.

\paragraph{Social biases.} Diffusion image generation  models have been found to be prone to social and cultural biases \cite{struppek2022biased, cho2022dalleval, luccioni2023stable}, even exaggerating and amplifying societal stereotypes beyond simple imbalances in the training data \cite{binachi2023, fraser2023friendly}, although quantifying amplification can be subtle \cite{seshadri-amplification-paradox}. Previous work has tackled this issue by modifying model parameters after training, by projecting out biased directions in the text embedding \cite{chuang2023debiasing}, or by performing algebraic manipulation of the representations \cite{wang2023concept}. One previous work, which inspires our current method, applies a direct closed-form model editing method \cite{orgad2023editing}.
The previous works have found that debiasing multiple concepts simultaneously is challenging, because debiasing one concept affects others, even in the presence of regularization methods. Our method overcomes that limitation with a new debiasing procedure that eliminates the mutual effect between concepts.

\paragraph{Model editing.} Model editing has recently emerged as an approach to control a model's behavior without training. In model editing, a subset of the model's weights is modified by locating the knowledge in the model and targeting it. Notably, \cite{bau2020rewriting} introduced a closed-form approach to manipulate specific rules encoded in deep generative vision models, enabling users to interactively modify these rules.  Closed-form solutions for editing knowlege in generative text models have been proposed in \cite{meng2022locating, meng2022mass}, while \cite{orgad2023editing, arad2023refact} have edited knowledge in text-to-image diffusion models by targeting either the cross-attention layers or the text-encoder layers.  Our method adopts and generalizes these approaches to enable removal and debiasing of many concepts simultaneously in text-to-image models.%

\section{Background}
    Diffusion models are generative models that can approximate distributions through a gradual denoising process \cite{sohl2015diffusion,ho2020denoising}. Starting from Gaussian noise, the model iteratively denoises over $T$ time steps to form a final image. At each intermediate step $t$, the model predicts noise $\epsilon_t$ that is added to the original image, with $x_T$ as initial noise and $x_0$ as the final output. By learning the parameters of the denoising process, the trained model can generate novel images from noise. This denoising is modeled as a Markov transition probability.
\begin{align}
    p_{\theta}(x_{T:0}) = p(x_T)\prod_{t=T}^{1}p_{\theta}(x_{t-1} | x_t)
\end{align}

Text-to-image latent diffusion models operate on low-dimensional embedding that is modeled with a U-Net generation network. 
The text conditioning is fed to the network via text embedding, extracted from a language model, in the cross-attention layers. Specifically, the attention modules within diffusion models follow the QKV (Query-Key-Value) \cite{vaswani2017attention} structure, where queries originate from the image space, while keys and values are derived from the text embeddings. Our focus centers on the linear layers $W_k$ and $W_v$, responsible for projecting text embeddings.

For a given text embedding $c_i$, the keys and values are generated by $k_i = W_kc_i$ and $v_i = W_vc_i$ respectively. The keys are then multiplied by the query $q_i$ that represents the visual features of the current intermediate image. This produces an attention map that aligns relevant text and image regions:
\begin{align}
    \mathcal{A} \propto \mathrm{softmax} (q_ik_i^T)
\end{align}

The attention map indicates the relevance between each text token and visual feature. Using this alignment, the cross-attention output is then computed by attending over the value vector V with the normalized attention weights.
\begin{align}
\label{eq:xattn_output}
    \mathcal{O} = \mathcal{A}v_i
\end{align}
The cross-attention is the mechanism that links the text and image information and responsible for assigning visual meaning to text tokens. The output of Equation~\ref{eq:xattn_output} is then propagated through the remaining layers of the diffusion U-Net. \par
\begin{figure}
    \centering
    \includegraphics[width=1\linewidth]{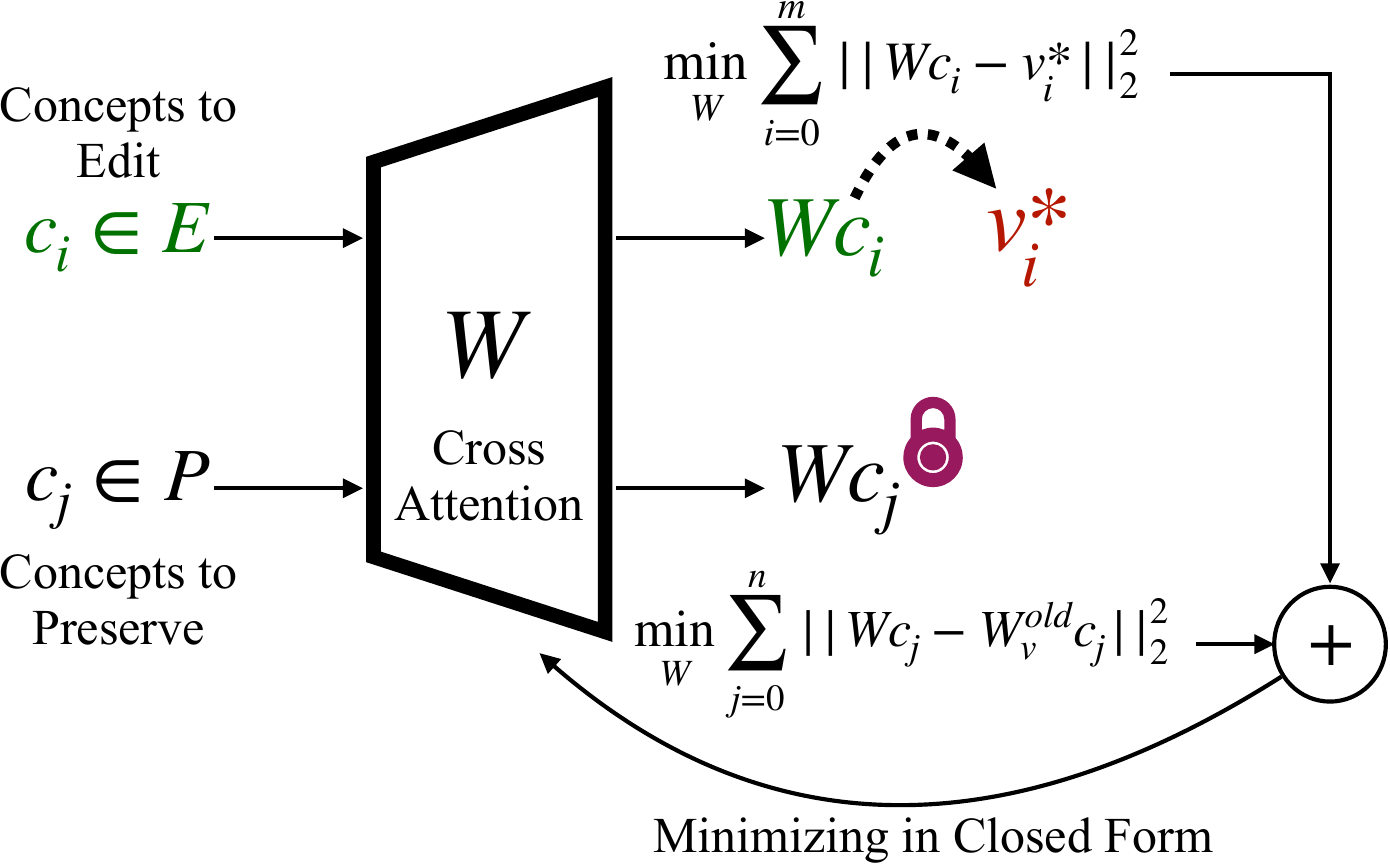}
    \caption{Closed-form editing of cross-attention weights enables concept manipulation in diffusion models. Our method modifies the attention weights  to induce targeted changes to the keys and values corresponding to specific text embeddings for a set of edited concepts $c_i\in E$ while minimizing changes to a set of preserved concepts $c_j\in P$. That dual objective allows debiasing, erasing, or moderating concepts while preserving unrelated ones. The same editing function is applied in all cases, but the target keys and values are set differently per application. As a closed-form edit, modifying attention weights given the new keys and values mappings takes less than 1 minute. That enables efficient simultaneous editing of multiple concepts.}
    \label{fig:method_figure}
\end{figure}

TIME \cite{orgad2023editing} edits implicit assumptions in pre-trained diffusion models by updating the cross-attention layers. Implicit assumptions can be any visual features that a model assumes about objects in an under-specified prompt, such as the color of roses which is usually red, or the gender of a doctor which is usually male. 
To edit these assumptions, the method requires a "source" under-specified prompt where the model makes an assumption (e.g. "a pack of roses") and a "destination" prompt specifying the desired attribute (e.g. "a pack of blue roses"). TIME updates the projection matrices $W_k$ and $W_v$, to bring the source prompt embedding closer to the destination embedding. This aligns the textual concepts such that the model no longer makes the implicit assumption.

Let $c_i$ be the source embedding, derived from the tokens of the source prompt, and $c_i*$ be the corresponding destination embeddings, taken from the embeddings of the corresponding tokens in destination prompt. The values of the destination prompts are calculated as $v_i* = W^\mathrm{old}c_i*$. New projection matrices $W$ are then optimized to minimize the objective function (a similar equation for the key projection matrices can be derived):

\begin{align}
    \min_{W} \sum_{i=0}^{m}||Wc_i - \underbrace{v_i^*}_{W^\mathrm{old}c_i^*}||_2^2 +\lambda ||W - W^\mathrm{old}||_F^2
\end{align}
where $\lambda$ is a regularization hyper-parameter. \cite{orgad2023editing} proved that the loss function has a closed-form global minimum solution, which allows efficient editing of text-to-image models.

\begin{align}
    W  = \left(\sum_{i=0}^{m} v_i^* c_i^T + \lambda W^\mathrm{old}\right) \left(\sum_{i=0}^{m} c_i c_i^T + \lambda \mathbb{I}\right)^{-1}
\end{align}

The first term in the inverse matrix, $\sum\limits_{i=0}^{m} c_i c_i^T$, is the covariance of the concept text embeddings being edited. As discussed in the appendix, we interpret the second term, an identity matrix, as matching the covariance of the large encyclopedia of concept embeddings in the diffusion model's vocabulary, inspired by MEMIT \cite{meng2022mass}. \par

While TIME formulation is effective, it risks interference with surrounding concepts when editing a particular concept. For example, editing doctors to be female might also affect teachers to be female. TIME has a regularization term that prevents the edited matrix from changing too radically. However, it is a general term and thus affects all vector representations equally. In this work, we present an alternative preservation term that allows targeted editing of the parameters of the pretrained generative model while maintaining its core capabilities.

\section{Method}

We introduce a general model editing methodology applicable to any linear projection layer. Given a pretrained layer $W^\mathrm{old}$, our goal, as shown in Figure~\ref{fig:method_figure}, is to find new edited weights $W$ that edit a set of concepts in set $E$ while preseving a set of concepts in set $P$.  Specifically, we wish to find weights so that the output for each of the inputs $c_i\in E$ maps to target values $v_i^* = W^\mathrm{old}_v c_{i^*}$ instead of the original $W^\mathrm{old}c_i$, while preserving outputs corresponding to the inputs $c_j\in P$ as $W^\mathrm{old}c_j$. A formal objective function can be constructed as:
\begin{align}
\label{eq:obj}
\min_{W} \sum_{c_i\in E}||Wc_i - v_i^*||_2^2 + \sum_{c_j\in P}||Wc_j - W^\mathrm{old}c_j||_2^2
\end{align}
As derived in the Appendix, the objective function in Equation~\ref{eq:obj} has a closed-form solution for the updated weights:
\begin{align}
    W  = \text{\footnotesize$\left(\sum_{c_i\in E}\hspace{-3pt} v_i^* c_i^T + \hspace{-3pt}\sum_{c_j\in P}\hspace{-3pt} W^\mathrm{old} c_jc_j^T\right)\left(\sum_{c_i\in E}\hspace{-3pt} c_i c_i^T + \hspace{-3pt}\sum_{c_j\in P}\hspace{-3pt} c_j c_j^T \right)^{-1}$}
    \label{eq:closed_form}
\end{align}
This formulation generalizes both the TIME\cite{orgad2023editing} and MEMIT\cite{meng2022mass} editing methods. When only canonical directions of the inputs are used as preservation terms $c_j$, our method reduces to TIME. Solving for the weight update $\Delta W$ instead of directly solving for $W$, our method reduces to MEMIT closed-form update. We discuss in detail how our approach provides a unified generalization that encompasses prior editing techniques as special case in the Appendix. \par

We edit the linear cross-attention projections ($W_k$ and $W_v$) to perform various concept edits with different goals: erasure, moderation, and debiasing. Our method requires the $m$ text embeddings $c_i$ derived from text descriptions of the concepts to edit and their corresponding modified target outputs $v_i^*$. The target outputs are defined differently based on the edit type, through the destination concepts $c_i^*$ as described below. We also preserve $n$ surrounding concepts using their descriptions $c_j$. For concepts with multiple tokens, we align the last token of $c_i$ to the last token of $v_i^*$ and make the edit.

\paragraph{Erasing} To erase a concept $c_i$, we want to prevent the model from generating it. If the concept is abstract like an artistic style (e.g. ``Kelly Mckernan''), this can be accomplished by modifying the weights so the target output $v_i$ aligns with a different concept $c_*$ (e.g. ``art''):
\begin{align}
    v_i^* \gets W^\mathrm{old}c_*
\end{align}
This updates the weights such that the output no longer reflects concept $c_i$, effectively erasing that concept from the model's generations and eliminating generations of the undesired characteristics.

\paragraph{Debiasing} To debias a concept $c_i$ (e.g. ``doctor'') across attributes $a_1, a_2, ..., a_p$ (e.g. ``white'', ``asian'', ``black'', ..), we want the model to generate the concept with evenly distributed attributes. This is achieved by adjusting the magnitude of $v_i$ along the directions of $v_{a_1}, v_{a_2}, ..., v_{a_p}$, where $v_{a_i} = W^\mathrm{old}a_i$ corresponds to the attribute text prompts:
\begin{align}
\label{eq:bias}
    v_i^* \gets W^\mathrm{old}\left[ c_i +\alpha_1 a_1 +\alpha_2 a_2 +...+\alpha_p a_p \right]
\end{align}
The constants $\alpha_i$ are chosen such that the diffusion model generates the concept with any desired probability for each attribute. This enables our method to debias multiple attributes simultaneously, unlike previous approaches such as TIME and concept ablation that can debias across dual attributes only. We provide the detailed algorithm in Alg \ref{algo:debias}.

\begin{algorithm}
\caption{Debiasing Concepts in Diffusion Models}
\label{algo:debias}
\begin{algorithmic}[1]
\State \textbf{Input:} Diffusion $M$ with cross attentions $W_k, W_v$
\State \textbf{Input:} Edit list $E$, preserve list $P$
\State \textbf{Input:} Attributes $A$ (list of strings of size $p$)
\State \textbf{Input:} Learning step $\eta$, desired ratios $R_{des}$

\While{True}
\State $R_{curr} \gets$ \Call{get\_ratios}{$M, E, A$}  
\For{$i, c_i$ \textbf{in} enumerate($E$)}
    \If{$max(|R_{curr}[i] - R_{des}[i]|) < 0.05$}
        \State $P$.append($c_i$) \Comment{add to preserve post debias}
        \State $E$.remove($c_i$) \Comment{remove from edit list}
        \State \textbf{continue}
    \EndIf
    \State $\alpha \gets \eta(R_{curr}[i] - R_{des}[i])$ \Comment{\text{$\alpha \in \mathcal{R}^p$}}
    \State $v^*_i \gets W_vc_i + \alpha \cdot A$  
    \State $k^*_i \gets W_kc_i + \alpha \cdot A$  
\EndFor
\If{$E$ is empty}
    \State \textbf{break} \Comment{All concepts debiased}
\EndIf
\State $W_v$  = \Call{uce}{$E$, $\{v_i^*\}$, $P$, $W_v$} \Comment{\Call{uce}{} is Eq.\ref{eq:closed_form}}%

\State $W_k$  = \Call{uce}{$E$, $\{k_i^*\}$, $P$, $W_k$}%
\EndWhile 
\State \textbf{return} $M$ \Comment{Debiased Model}
\end{algorithmic}
\end{algorithm}

\paragraph{Moderation} To moderate concept $c_i$ (e.g. ``nudity''), we perform an edit where the target output $v_i^*$ aligns with an unconditional prompt $c_0$ (e.g. `` ''):
\begin{align}
    v_i^* \gets W^\mathrm{old}c_0
\end{align}
This replaces the output for $c_i$ with a more generic, unconditional output $c_0$, moderating the model's response by reducing extreme attributes of that concept.

\section{Experiments}
    \subsection{Erasing}

Our erasing technique directly modifies the key--value mappings in the model to associate keys with different concepts rather than the undesired ones. We use our method to erase artistic styles from the model's weights. Our technique allows preserving certain artists while removing others. We found this enables substantially less interference on a holdout set of artists that were neither erased nor explicitly preserved.
 We compare our artistic erasure method to recent approaches including ESD-x\cite{gandikota2023erasing}, Concept Ablation \cite{kumari2023ablating}, and SDD\cite{kim2023towards} which use cross-attention fine-tuning for controllable image editing.
 In a second set of experiments, we test object erasure (e.g., erasing the concept of garbage trucks). In this set of experiments, we did not use any explicit preservation objectives, in order to test implicit interference.
 For object erasure, we primarily compare to ESD-u\cite{gandikota2023erasing}, which freezes all parameters except cross-attentions during fine-tuning, enabling more global erasures. 
\subsubsection{Artist erasure}
\begin{figure}
    \centering
    \includegraphics[width=\linewidth]{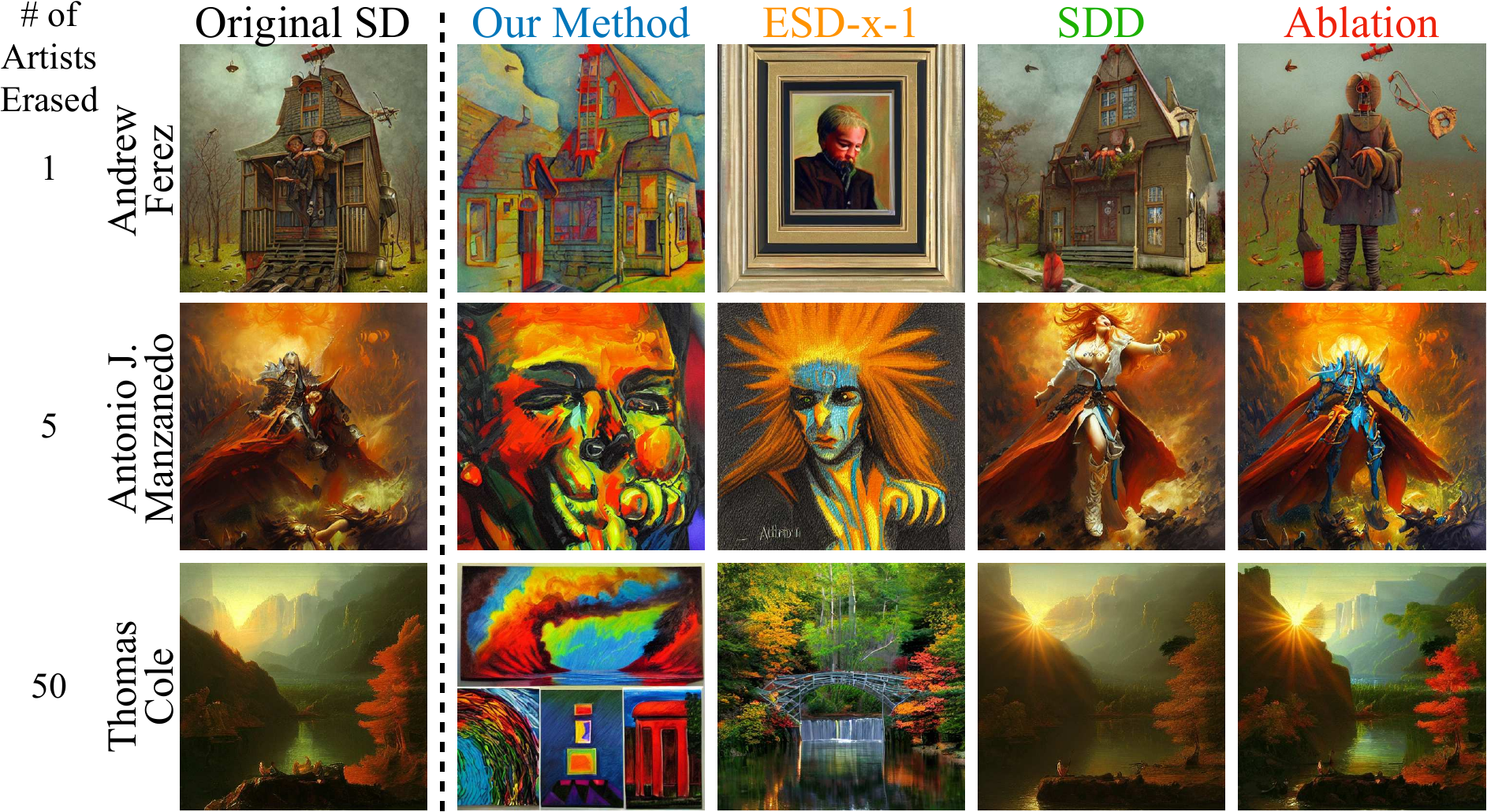}
    \caption{Our method and ESD-x show strong erasing capabilities. SDD and Ablation\protect\footnotemark start to dilute their erasing capabilities as the number of concepts being erased are increased.}
    \label{fig:intended}
\end{figure}

Our method can successfully erase multiple concepts while preserving the model's knowledge. We use the text embeddings of the artist names as our concepts $c_i$ to erase and a set of artists to preserve $c_j$. As shown in Figure \ref{fig:intended}, we are able to consistently erase multiple artistic styles, while other methods maintain a lot of characteristics of the artistic styles and impair the model's capabilities as the number of erased concepts increases. We found ESD and SDD tend to damage the model more when erased sequentially (at 1000 iterations per concept), so we limited those techniques to random sampling-based erasure for a fixed 1000 iterations.\footnote{The authors of SDD note potential overfitting of their method when erasing multiple concepts. To mitigate this, we limited SDD to 700 iterations for multi-concept erasure. For Ablation, the authors suggest 100 iterations per concept, however we found the model deteriorates after 1000 total iterations. Therefore, we restricted Ablation to 1000 iterations total when erasing multiple concepts.}\par

\begin{figure*}
    \centering
    \includegraphics[width=\linewidth]{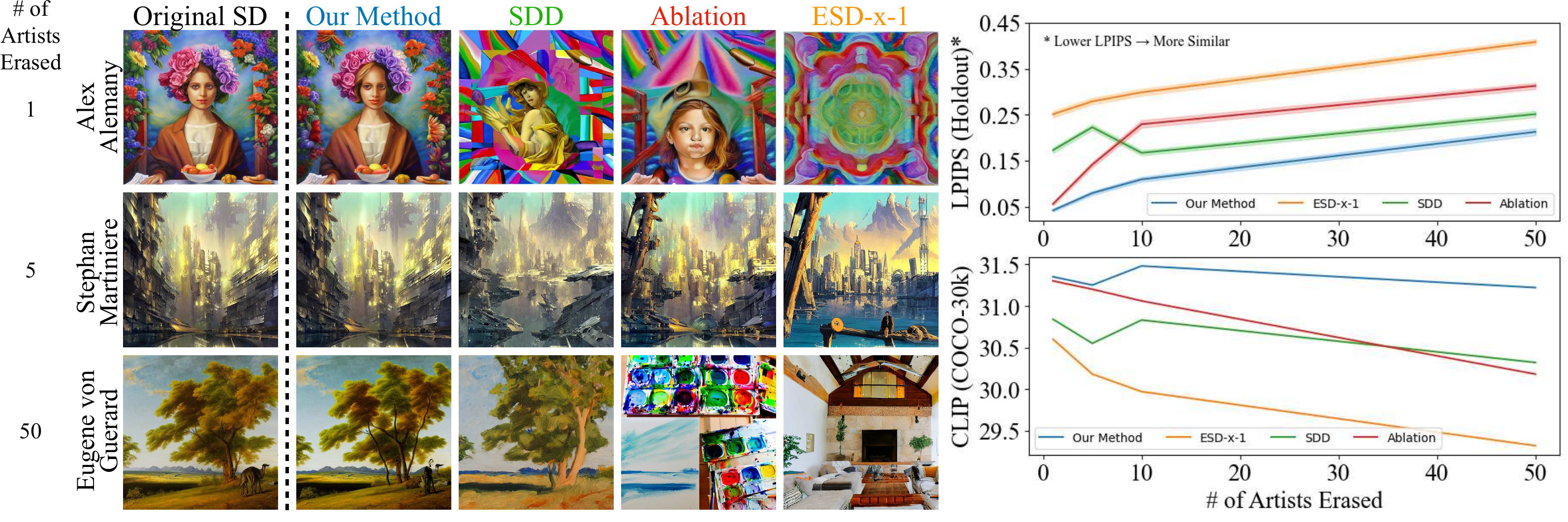}
    \caption{Our method preserves the remaining knowledge of the model better after the edit. The figure shows images generated from different editing methods, for the same prompts and seeds, across a variety of \textbf{artists that are not erased}. Our method exhibits lower LPIPS, indicating less change to unerased concepts during model editing. Similarly for COCO, we find that our method has better CLIP scores across all the scales. This demonstrates that our method has significantly reduced interference compared to other fine-tuning approaches when editing.}
    \label{fig:interference}
\end{figure*}

Our method also demonstrates reduced interference with neighboring, non-erased concepts compared to other techniques. As shown in Figure \ref{fig:interference}, erasing with our approach has less impact on concepts that were not targeted for removal. The top plot shows the LPIPS\cite{zhang2018unreasonable} difference between the original SD and edited models, indicating our method results in the smallest modifies the unrelated concepts the least. The bottom plot shows the CLIP score\cite{radford2021learning} on COCO-30k prompts~\cite{lin2014microsoft}, where our method maintains better text-to-image alignment after editing, as evidenced by the higher CLIP score. Together, these results highlight an important advantage of our erasing approach -- the ability to remove targeted concepts with minimal disruption to other areas of knowledge in the model. \par
Diffusion models were shown to mimic more than 1800 artistic styles \cite{sd2022artists}. We analyzed the capabilities of our method to erase multiple concepts by erasing $n$ artists while preserving the remaining $1000-n$. As shown in Table \ref{tab:limit_coco}, our method can erase up to 100 artists simultaneously before damaging image fidelity and CLIP scores. After 50 erasures, the model's output for a given prompt and seed begins to change , as indicated by the LPIPS score, but remains aligned overall as evidenced by the CLIP score. The importance of our preservation strategy to these results is shown in the Appendix, where no preservation reduces back to the TIME formulation. \par

\begin{table}
    \centering
    \begin{tabular}{lccc}
        \textbf{\# Concepts} & \textbf{CLIP $\uparrow$}  & \textbf{LPIPS$\downarrow$} & \textbf{FID$\downarrow$} \tabularnewline
         \hline
         1 & 31.35 & 0.05 & 14.37 \tabularnewline
         5  & 31.25 & 0.08 & 14.30 \tabularnewline
         10  & 31.48 & 0.13 & 15.56  \tabularnewline
         50 & 31.22 & 0.22 &  15.10 \tabularnewline
         100  & 30.08 & 0.30 &  15.09 \tabularnewline
         500  & 21.06 & 0.44 & 72.40 \tabularnewline
         1000 & 16.79 & 0.47 & 85.48 \tabularnewline
         \cdashline{1-4}
         Original SD & 31.32 & - & 14.49 \tabularnewline
    \end{tabular}
    \caption{Our method can erase upto 100 concepts while performing similar to pre-trained SD on COCO-30k dataset. The image fidelity is consistent with original SD till 100 erasures. With LPIPS, we find that the model at 100 erasures has a slightly different performance for a given seed and prompt, but as the CLIP score shows, the alignment of the model is still intact. }
    \label{tab:limit_coco}
\end{table}

\subsubsection{Erasing Objects}
To demonstrate the capability of our method to erase objects from the diffusion model's learned concepts, with potential applications for removing harmful symbols and content, we conducted experiments erasing Imagenette~\cite{howard2020fastai} classes, a subset of Imagenet classes~\cite{deng2009imagenet}. For each erased object, we utilized the text embedding (e.g. "French Horn") as $c_i$, without additional preservation concepts $c_j$. We generated 500 images per class and evaluated top-1 classification accuracy using a pretrained ResNet-50\cite{he2016deep}, comparing to ESD-u in Table \ref{tab:imagenette}. Objects were erased individually to analyze interference versus ESD on non-erased classes. Without explicit preservation, our approach exhibited superior erasure capability while minimizing interference on non-targeted classes. Further erasure analysis is provided in the Appendix. Erasing all 10 Imagenette classes together reduced image generation accuracy to just 4.0\% and COCO-CLIP score to 31.02 (original SD is 31.32), quantitatively showing effective single and multi-object removal while limiting interference.
\begin{table}
    \centering
    \resizebox{\columnwidth}{!}{
    \begin{tabular}{@{\extracolsep{4pt}}lcccccc@{}}%
        \textbf{Class name} & \multicolumn{3}{C{3cm}}{\textbf{Accuracy of Erased Class $\downarrow$}} & \multicolumn{3}{C{3cm}}{\textbf{Accuracy of Other Classes $\uparrow$}}
        \tabularnewline
        \cline{2-4}\cline{5-7} 
         & \textbf{SD} & \textbf{Ours} &\textbf{ ESD-u }& \textbf{SD} & \textbf{Ours} & \textbf{ESD-u} \tabularnewline
        \hline 
        Cassette Player & 15.6 & 0.0  & 0.60 & 85.1 & 90.3   & 64.5\tabularnewline
        Chain Saw & 66.0 & 0.0  &  6.0 & 79.6 &  76.1 &  68.2\tabularnewline
        Church & 73.8 & 8.4  &  54.2 & 78.7 &  80.2  & 71.6\tabularnewline
        Gas Pump & 75.4 &  0.0 &  8.6 & 78.5 &  80.7  & 66.5\tabularnewline
        Tench & 78.4 &  0.0 &  9.6 & 78.2 & 79.3  &  66.6\tabularnewline
        Garbage Truck & 85.4 & 14.8  &  10.4 & 77.4 &  78.7 &  51.5\tabularnewline
        English Springer & 92.5 & 0.2  &  6.2 & 76.6 &  78.9 &  62.6\tabularnewline
        Golf Ball & 97.4 & 0.8  &  5.8 & 76.1 &  79.0 &  65.6\tabularnewline
        Parachute & 98.0 & 1.4  &  23.8 & 76.0 & 77.4  &  65.4\tabularnewline
        French Horn & 99.6 & 0.0  &  0.4 & 75.8 &  77.0 &  49.4\tabularnewline
        \cdashline{1-7} 
        Average & 78.2 & \textbf{2.6}  &  12.6 & 78.2 & \textbf{79.8}  &  63.2
    \end{tabular}}
    \caption{Our method can erase objects from diffusion models effectively without impacting the accuracy for other object classes even when they are not explicitly preserved. Compared to \texttt{ESD-u}, we demonstrate improved erasure of the targeted class alongside higher preservation of unrelated classes in the generated images on Imagenette classes.}
    \label{tab:imagenette}
    \vspace{-1em}
\end{table}

    \subsection{Debiasing}
Stable Diffusion exhibits gender and racial bias when generating images for profession names (e.g. CEO), producing only 6\% female figures for "CEO" prompt. We debias profession concepts via Alg. \ref{algo:debias}, using profession text embeddings $c_i$ and attribute embeddings $A$ (e.g. "male", "female"). To prevent over/under-debiasing, we set per-attribute regularization constants $\alpha_i$ in Eq. \ref{eq:bias}. As debiasing one concept can affect others\cite{orgad2023editing}, we use an iterative approach. We maintain edit and freeze concept lists, fixing debiased concepts while editing new ones. With multiple concepts debiased in parallel, $\alpha_i$ values are found by generating validation samples during training and adjusting constants based on the current model's generated ratio (classified by CLIP). Once a concept is sufficiently debiased, we add it to a preservation list, bypassing validation and keeping it fixed when debiasing others. This iterative $\alpha_i$ tuning enables efficient debiasing by avoiding unnecessarily repeated editing of already debiased concepts. Setting equal $\alpha_i$ for all concepts risks over-debiasing some while under-debiasing others. Our iterative validation determines optimal per-concept constants. %

\subsubsection{Gender bias}

Prior methods for debiasing generative models like TIME\cite{orgad2023editing}, Concept Algebra \cite{wang2023concept}, and Debiasing-VL\cite{chuang2023debiasing} have focused on mitigating biases between two discrete attributes. While we acknowledge that a binary perspective of gender excludes non-binary groups, for a fair comparison to such dual-attribute techniques, we evaluate our method by reducing occupational gender biases in diffusion models. We recognize that editing for visual features of non-binary genders risks introducing other unwanted stereotypical behavior.

\begin{figure}
    \centering
    \includegraphics[width=\linewidth]{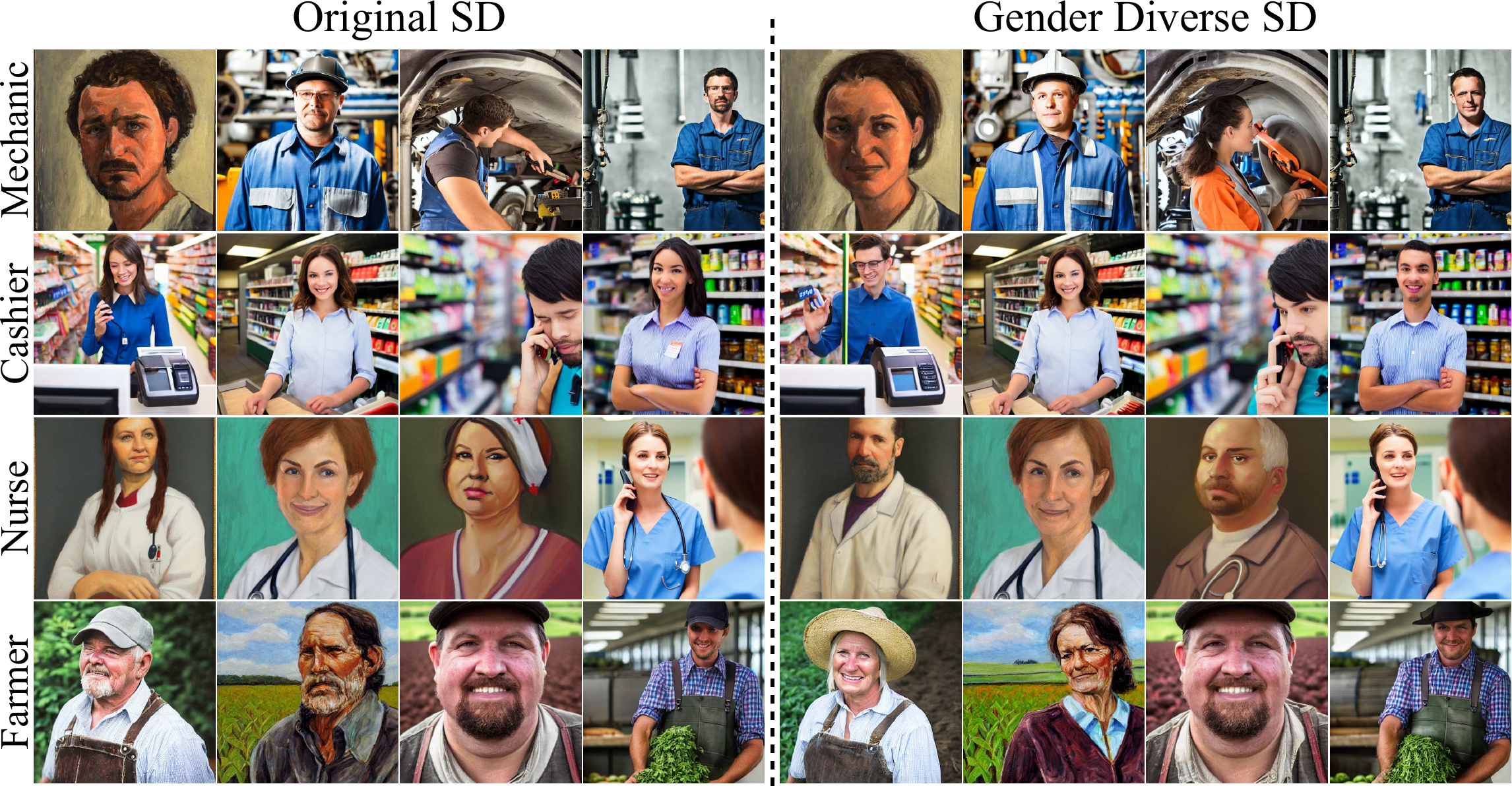}
    \caption{Our method improves the gender representation of professions in the stable diffusion generated images. We find that the images precisely change the gender while keeping the rest of the scene intact.}
    \label{fig:gender}
\end{figure}

Figure~\ref{fig:gender} provides qualitative results demonstrating increased diversity in generated images for professions with strong initial gender biases after applying our proposed debiasing technique. For quantitative evaluation, we synthesize 250 images per profession and utilize CLIP classifications to calculate the deviation $\Delta = \frac{|p_{\text{desired}} - p_{\text{actual}}|}{p_{\text{desired}}}$ between the achieved and desired (50-50) gender ratios, where $\Delta=0$ indicates perfect debiasing. As shown in Table~\ref{tab:gender_bias}, our method achieves gender distributions closest to the balanced 50-50 ratio compared to pretrained and baseline models. The original formulation of TIME~\cite{orgad2023editing} exhibits interference between debiased concepts, resulting in worse performance. We find that even when applying TIME with our proposed preservation term, it still underperforms compared to our approach. Through both qualitative and quantitative results, we demonstrate that our method enables robust targeted debiasing of generative models.

\begin{table*}
    \centering
    \begin{tabular}{lcccccc}
        \textbf{Profession} & \textbf{Original-SD}  & \textbf{Concept Algebra} & \textbf{Debias-VL} & \textbf{TIME} & \textbf{TIME + Preserve} & \textbf{Ours} \tabularnewline
        
         \hline
         Librarian & 0.86 $\pm$ 0.06 & 0.66 $\pm$ 0.07 & 0.34 $\pm$ 0.06 & 0.26 $\pm$ 0.05 & 0.35 $\pm$ 0.01 & \textbf{0.07 $\pm$ 0.07} \tabularnewline
         Teacher & 0.42 $\pm$ 0.01 & 0.46 $\pm$ 0.00 & 0.11 $\pm$ 0.05 & 0.34
 $\pm$ 0.06 & 0.07 $\pm$ 0.06 & \textbf{0.06 $\pm$ 0.02} \tabularnewline
         Sheriff & 0.99 $\pm$ 0.01 & 0.38 $\pm$ 0.22 & 0.82 $\pm$ 0.08 & 0.22 $\pm$ 0.05 & 0.10 $\pm$ 0.05 & \textbf{0.10 $\pm$ 0.03} \tabularnewline
         Analyst & 0.58 $\pm$ 0.12 & 0.24 $\pm$ 0.18 & 0.71 $\pm$ 0.02 & 0.52 $\pm$ 0.03 & \textbf{0.13 $\pm$ 0.05} & 0.20 $\pm$ 0.07 \tabularnewline
         Doctor & 0.78 $\pm$ 0.04 & 0.40 $\pm$ 0.02 & 0.50 $\pm$ 0.04 & 0.58 $\pm$ 0.03 & 0.41 $\pm$ 0.08 & \textbf{0.20 $\pm$ 0.02} \tabularnewline
         \cdashline{1-7}
         WinoBias \cite{zhao2018gender} & 0.67 $\pm$ 0.01 & 0.43 $\pm$ 0.01 & 0.55 $\pm$ 0.01 & 0.44 $\pm$ 0.00 & 0.31 $\pm$ 0.00 & \textbf{0.22 $\pm$ 0.00} \tabularnewline
    \end{tabular}
    \caption{Debiasing performance on 5 randomly-picked professions and an average on all 35 Winobias\cite{zhao2018gender} professions. The presented metric $\Delta$ measures the percentage deviation from desired ratios ($\Delta=0$ indicates complete debiasing). Our method has a consistent debiasing performance compared to previous inference and model editing methods by showing the least average deviation from the desired distribution.}
    \label{tab:gender_bias}
\end{table*}

\subsubsection{Racial bias}
\begin{figure}
    \centering
    \includegraphics[width=\linewidth]{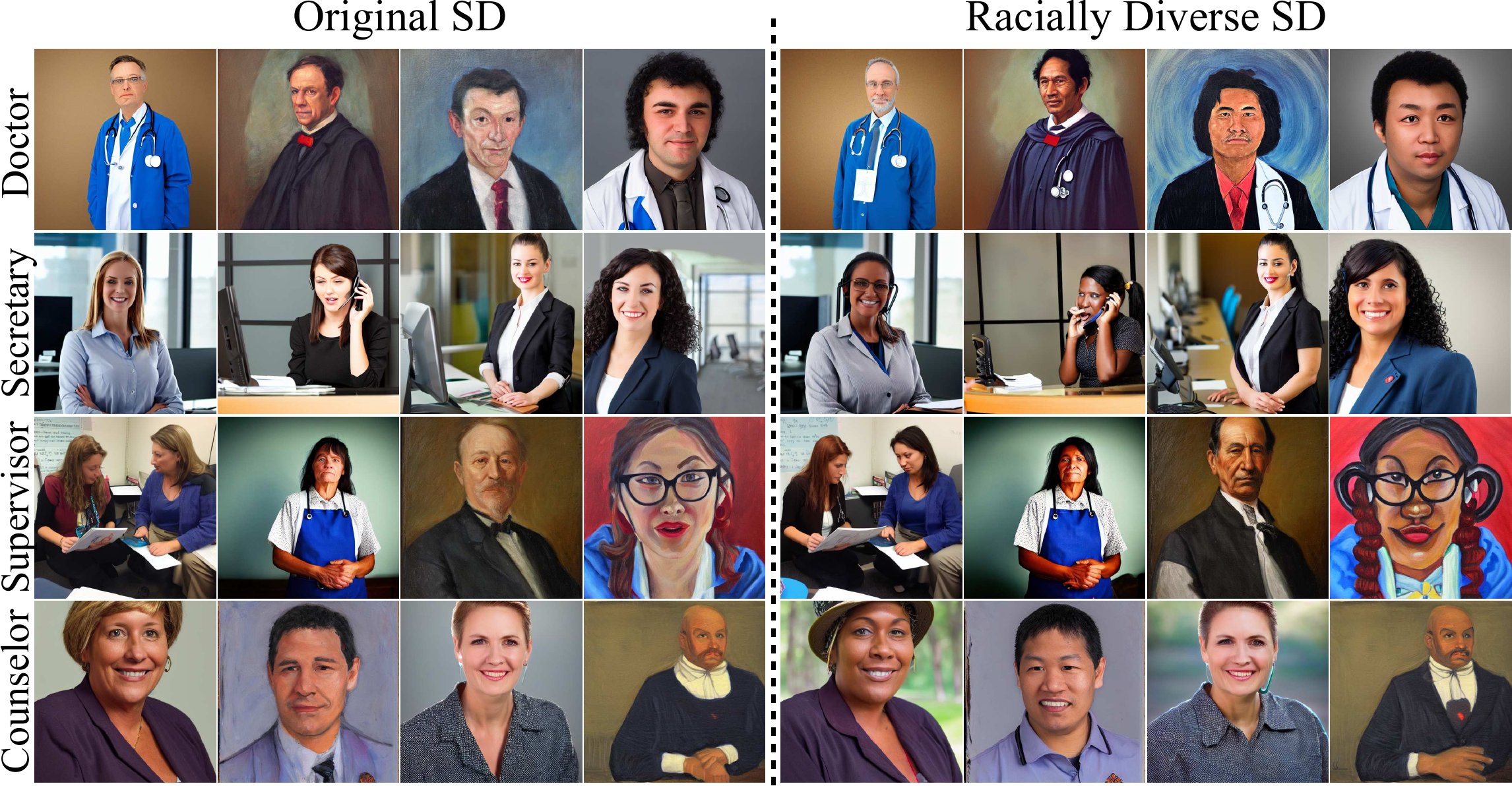}
    \caption{Our method improves the racial diversity of professions in the pre-trained stable diffusion. We show images from the original SD and the corresponding images from the edited model for the same prompts and seeds for comparison. We find that our edited model has a better race representation.}
    \label{fig:race}
\end{figure}

A key advantage of our approach over prior debiasing techniques is the ability to concurrently mitigate biases related to multiple attributes. To demonstrate this capability, we conduct experiments to improve racial diversity in professions generated by Stable Diffusion. Specifically, we target major racial categories as defined by U.S. Office of Management and Budget (OMB) standards\cite{OMB2022standards}: White, Black, American Indian, Native American, and Asian. Accurately classifying race from images is an intricate task, problematic even for sophisticated models like CLIP and humans. We, therefore, take a qualitative analysis approach rather than attempting error-prone quantitative race categorization. As depicted in Figure \ref{fig:race}, our method significantly enhances the representation of these racial groups among generated professional images. This highlights our technique's strength in reducing multifaceted biases in diffusion models, a key advantage over existing binary-attribute debiasing methods.

    \subsection{Moderation}

\begin{figure}
    \centering
    \includegraphics[width=1\linewidth]{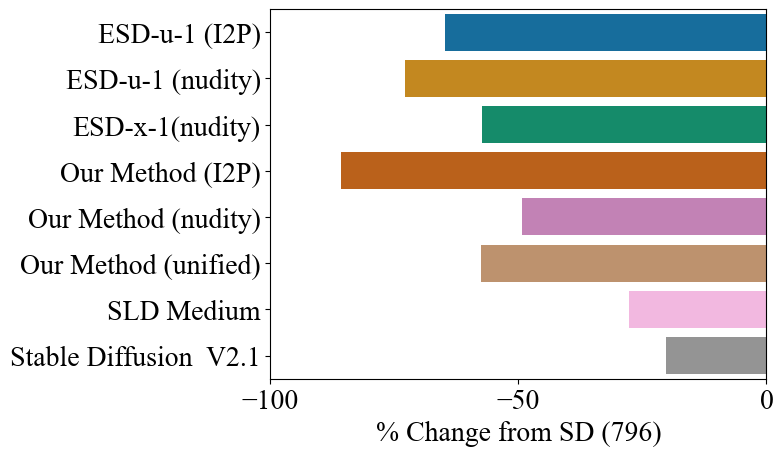}
    \caption{A percentage reduction in nudity-classified samples on I2P prompts compared to SD.
    Our method erases nudity content from pre-trained SD and has the advantage of erasing multiple concepts in I2P prompts. \texttt{"Nudity"} erased model performs very similar to \texttt{ESD-x-1} as both the methods edit only cross attentions. However, as noted in Table~\ref{tab:moderation_fid}, we find that our method results in a finer edit and has better alignment with COCO.}
    \label{fig:nudity_bar}
\end{figure}

We quantitatively evaluate our proposed method for moderating sensitive concepts, comparing it against recent state-of-the-art techniques ESD-u and ESD-x\cite{gandikota2023erasing} on the task of erasing single concepts like "nudity". For all the models, 4703 images are generated using the prompts from the I2P benchmark introduced in \cite{schramowski2023safe}. In Figure \ref{fig:nudity_bar} we analyze the nudity moderation using NudeNet classifier\cite{bedapudi2022nudenet}. We find that our method demonstrates comparable nudity erasure performance to ESD-X since both techniques edit cross-attentions of the diffusion model. ESD-u as expected has a more aggressive erasure effect given it finetunes the entire model except cross attentions. However, Table \ref{tab:moderation_fid} highlights that our approach induces substantially lower distortion to model generations than ESD-u and ESD-x, with significantly reduced LPIPS\cite{zhang2018unreasonable} score    from the original SD generations. This indicates our method better preserves image quality while moderating sensitive concepts. Additionally, the CLIP score indicates that our technique maintains better text-image alignment post editing. \par
We further demonstrate efficacy in erasing multiple sensitive concepts from I2P \footnote{including hate, harassment, violence, suffering, humiliation, harm, suicide, sexual, nudity, bodily fluids, blood, obscene gestures, illegal activity, drug use, theft, vandalism, weapons, child abuse, brutality, cruelty}. Again, our approach shows improved multi-concept moderation capability compared to ESD-u (Figure \ref{fig:nudity_bar}). We provide a detailed analysis of moderating diverse sensitive concepts in the Appendix.

\begin{table}
    \centering
    \begin{tabular}{lcccc}
        \textbf{ Method} & \textbf{FID-Real $\downarrow$}  & \textbf{FID-SD $\downarrow$} & \textbf{CLIP $\uparrow$} & \textbf{LPIPS $\downarrow$}\tabularnewline
         \hline
         REAL & - & 14.49 & 30.41 & - \tabularnewline
         SD  & 14.49 & - & 31.32 &  - \tabularnewline
         ESD-u-1 & \textbf{14.16} & 3.73 & 30.45 & 0.23\tabularnewline
         ESD-x-1 & 14.45 & 2.33 & 30.81 & 0.18\tabularnewline
         Ours  & 14.84 &\textbf{ 1.82 }& \textbf{31.26} & \textbf{0.12}\tabularnewline
    \end{tabular}
    \caption{Our method performs  comparably to the pre-trained SD on COCO. The image fidelity performance compared to SD (FID-SD) and LPIPS matches closely with our method. FID with real COCO images (FID-real) is very similar to SD. Our method also has the closest CLIP score to the original SD compared to other methods.}
    \label{tab:moderation_fid}
    \vspace{-1em}
\end{table}

    \subsection{Unified Editing}
Our formulation enables simultaneous style erasure, profession debiasing, and nudity moderation. The edit vector $v^*$ design differs for each edit, but the model update is unified. Empirically, the jointly finetuned model demonstrates effectiveness on par with individually trained models: similar to Table~\ref{tab:gender_bias}, a gender ratio deviation of 0.27 versus 0.22 for the gender-debiasing model and 0.67 for the original Stable Diffusion. The unified model also shows a 58\% nudity reduction compared to 49\% for nudity erasure and 64\% for ESD-u, shown in Figure~\ref{fig:nudity_bar}.

\section{Limitations}
    When debiasing across multiple attributes, we find interdependencies that exhibit compounding biases. For example, generating images of "a black person" has near equal gender ratios (48\% male out of 100 images), while "a native american person" displays strong male bias (96\% male of 100). Debiasing in isolation can thus perpetuate biases along other dimensions. This highlights the need for joint attribute consideration to mitigate propagated biases. We also find word-level biases in prompts that compose unfavorably. Non-gendered phrases like "successful person" become predominantly male (88\% of 100) versus the gender-balanced "person" (50\% male of 100), illustrating how subtle cues carry biases. Such compositional effects pose challenges, as each word element contributes biases needing mitigation.\par
For artistic style erasure, removing over 500 artists degrades general image generation, even with preservation terms (Table~\ref{tab:limit_coco}). That suggests a critical mass of artists is needed to maintain generative capabilities. Excessive erasure impairs the core visual priors learned during pretraining.

\section{Conclusion}

We have presented a unified algorithm for precisely editing diffusion models to allow designers to make them more responsible and beneficial for society. 
Our approach enables targeted debiasing, erasure of potentially copyrighted content, and moderation of offensive concepts, using only text descriptions. Our measurements suggest that our method offers three key benefits over prior methods. First, it can mitigate multifaceted gender, racial, and other biases simultaneously while preserving model capabilities. Second, it is scalable, modifying hundreds of concepts in one pass without expensive retraining. Third, extensive experiments demonstrate superior performance on real-world use cases.
Together, our findings suggest that UCE is significant step towards democratizing access to ethical and socially-responsible generative models.
The ability to seamlessly unify debiasing, erasure, and moderation will be an important tool for building AI that benefits our diverse global society.

\section*{Acknowledgments}
Thanks to Antonio Torralba for valuable advice, discussions, and support.  RG and DB are supported by Open Philanthropy.
HO and YB are supported by the Israel Science Foundation (grant No.\ 448/20), Open Philanthropy, and an Azrieli Foundation Early Career Faculty Fellowship. The Center for AI Safety made computing resources available for this research.
{\small
\bibliographystyle{ieee_fullname}
\bibliography{mainbib}
}
\clearpage
\newpage
\mbox{}

\appendix
\counterwithin{figure}{section}
\counterwithin{table}{section}
\counterwithin{equation}{section}

\section{Deriving the Closed Form Solution for UCE}
Let $W_k$ and $W_v$ be the cross attention weights that project the text embeddings to image space corresponding to keys and values respectively. These are computed fresh at every time step making the computation straightforward, unlike the self-attentions.\par
Let $\{{c_i}\}_{i=0}^{m}$ be the embeddings of the text descriptions for the concepts we want to edit. Let $\{{c_j}\}_{i=0}^{n}$ be the embeddings of the concepts we wish to preserve. Similarly, let $\{{v_i^*}\}_{i=0}^{m}$ be the target cross-attention outputs that we want the concepts ${c_i}$ to be steered towards. The sets ${c_i}$, ${v_i^*}$, and ${c_j}$ represent the concepts to erase, desired target outputs, and concepts to preserve respectively. 
We optimize the newly edited weights of the cross-attention value projects $W$ by minimizing the loss function. The same optimization can be adopted to value projection optimization:
\begin{align}
\label{obj}
  \mathcal{L} = \sum_{c_i\in E}||Wc_i - v_i^*||_2^2+\sum_{c_j\in P}||Wc_j - W^{old}c_j||_2^2
\end{align}

The objective function in Equation~\ref{obj} can be solved to arrive at a closed-form solution. We take the derivative of the loss function stated above w.r.t to $W$ and set it to $0$.
\begin{align*}
    \sum_{c_i\in E}2(Wc_i - v_i^*)c_i^T+\sum_{c_j\in P}2(Wc_j - W^{old} c_j)c_j^T = 0
\end{align*}
\begin{align*}
    \hspace{-3pt}\sum_{c_i\in E}\hspace{-3pt}W c_i c_i^T - \hspace{-3pt}\sum_{c_i\in E}\hspace{-3pt} v_i^* c_i^T+ \hspace{-3pt}\sum_{c_j\in P}\hspace{-3pt} W c_j c_j^T - \hspace{-3pt}\sum_{c_j\in P}\hspace{-3pt} W^{old} c_jc_j^T = 0
\end{align*}
    
\begin{align*}
   \hspace{-3pt}\sum_{c_i\in E}\hspace{-3pt} W c_i c_i^T  + \hspace{-3pt}\sum_{c_j\in P}\hspace{-3pt} W c_j c_j^T = \hspace{-3pt}\sum_{c_i\in E}\hspace{-3pt} v_i^* c_i^T +  \hspace{-3pt}\sum_{c_j\in P}\hspace{-3pt} W^{old} c_jc_j^T 
\end{align*}
\begin{align*}
   W \hspace{-2pt}\left( \hspace{-1pt}\sum_{c_i\in E}\hspace{-3pt}c_i c_i^T  + \hspace{-3pt}\sum_{c_j\in P}\hspace{-3pt} c_j c_j^T \right) \hspace{-1pt} = \hspace{-2pt} \left( \hspace{-1pt}\sum_{c_i\in E}\hspace{-3pt} v_i^* c_i^T + \hspace{-3pt}\sum_{c_j\in P}\hspace{-3pt} W^{old} c_jc_j^T \right)
\end{align*}

To invert the terms on the left-hand side ($\sum_{c_i\in E}c_i c_i^T + \sum_{c_j\in P} c_j c_j^T$), the matrix must have full rank. Adding a preservation term increases the rank by 1. Thus if the number of preservation terms $|P| < d$, where $d$ is the dimension of the text embedding space, the matrix may not have full rank. To ensure the rank condition is satisfied, we introduce $d$ additional preservation terms along the canonical basis directions of the text embedding space. This maintains full rank and enables inversion of the matrix irrespective of the size of $P$.

\begin{align*}
    W = \footnotesize\text{$\left(\hspace{-1pt}\sum_{c_i\in E}\hspace{-3pt} v_i^* c_i^T + \hspace{-3pt}\sum_{c_j\in P}\hspace{-3pt} W^{old} c_jc_j^T\right)
    \left( \hspace{-1pt}\sum_{c_i\in E}c_i c_i^T  + \hspace{-3pt}\sum_{c_j\in P}\hspace{-3pt} c_j c_j^T\right)^{-1}$}
\end{align*}
We optimize both the cross-attention key and value weights using the same principles.

\section{UCE Generalizes to TIME}
Our method, Unified Concept Editing (UCE),  can be viewed as a generalization of the TIME method. As discussed in the methodology section, TIME regularizes the cross-attention weights. With our method, if we do not preserve any specific concepts and only preserve the canonical directions $e_j$ scaled by $\lambda$, we get the following closed-form solution:
\begin{align*}
     W \hspace{-1.5pt}= \hspace{-1.5pt}\footnotesize\text{$\left(\hspace{-1pt}\sum_{c_i\in E}\hspace{-3pt} v_i^* c_i^T + \lambda \hspace{-1pt}\sum_{j=0}^{d} W^{old} e_je_j^T\right)
    \hspace{-2pt}\left( \hspace{-1pt}\sum_{c_i\in E}\hspace{-3pt}c_i c_i^T  + \lambda\hspace{-1pt}\sum_{j=0}^{d} e_j e_j^T\right)^{-1}$}
\end{align*}
Where the canonical directions $e_j$ have outer products $e_ie_j^T$ that are diagonal matrices with only the $j^{th}$ element as 1 and rest 0. Summing all the canonical outer products gives the identity matrix $\mathbb{I}$.
\begin{align*}
     W = \footnotesize\text{$\left(\hspace{-1pt}\sum_{c_i\in E}\hspace{-3pt} v_i^* c_i^T + \lambda W^{old}.\mathbb{I}\right)
    \left( \hspace{-1pt}\sum_{c_i\in E}\hspace{-3pt}c_i c_i^T  + \lambda\mathbb{I}\right)^{-1}$}
\end{align*}
This is the closed-form solution for TIME, which regularizes equally across all directions. Our method can be seen as a generalization of TIME that adds preservation across important surrounding concepts, not just the canonical directions. This new formulation with additional explicit preservation is very practical, allowing us to edit multiple concepts with less interference. Our method builds on TIME by allowing the preservation of concepts beyond just the canonical directions. This helps enable editing multiple concepts with less interference.

\section{UCE Generalizes to MEMIT}
Our method can also be viewed as a generalization of MEMIT. Starting from our objective function in Equation~\ref{obj}:
\begin{align*}
  \mathcal{L} = \sum_{c_i\in E}||Wc_i - v_i^*||_2^2+\sum_{c_j\in P}||Wc_j - W^{old}c_j||_2^2
\end{align*}
Taking the derivative and setting it to zero gives:
\begin{align*}
    \sum_{c_i\in E}2(Wc_i - v_i^*)c_i^T+\sum_{c_j\in P}2(Wc_j - W^{old} c_j)c_j^T = 0
\end{align*}
To ensure full rank, instead of adding canonical directions to complete the rank, we add additional preservations for a plethora of concepts in diffusion vocabulary. We rewrite the equation in a block form by defining $v_j = W^{old} c_j$ and redefining $W$ as $W^{old} + \Delta W$:
\begin{align*}
    (W^{old} + \Delta W)(C_iC_i^T + C_jC_j^T) = V_iC_i^T + V_jC_j^T
\end{align*}
\begin{align}
    W^{old}C_iC_i^T + W^{old}C_jC_j^T + \Delta WC_iC_i^T + \Delta WC_jC_j^T  \nonumber \\ = V_iC_i^T + V_jC_j^T
\label{memit_expanded}
\end{align}
Assuming the preservation list contains most concepts the diffusion model knows $W^{old}$ will minimize $||Wc_j-v_j||_2^2$. Taking a derivative and equating to zero, we get $W^{old}C_jC_j^T=V_jC_j^T$. Subtracting this from Equation~\ref{memit_expanded} gives MEMIT closed form solution:
\begin{align*}
     \Delta W(C_iC_i^T + C_jC_j^T) = V_iC_i^T - W^{old}C_iC_i^T 
\end{align*}
With $R = V_i-W^{old}C_i$ and $C_0 = C_jC_j^T$
\begin{align*}
     \Delta W = RC_i^T (C_iC_i^T + C_jC_j^T)^{-1}
\end{align*}
In summary, our method generalizes MEMIT by incorporating additional preservation terms from the diffusion model's vocabulary and solving for the weight update $\Delta W$ instead of directly solving for $W$. This highlights the connection between our approach and existing techniques like MEMIT.

\section{Extended Experimental Results}

\subsection{Erasing Style}
\begin{figure}
    \centering
    \includegraphics[width=1\linewidth]{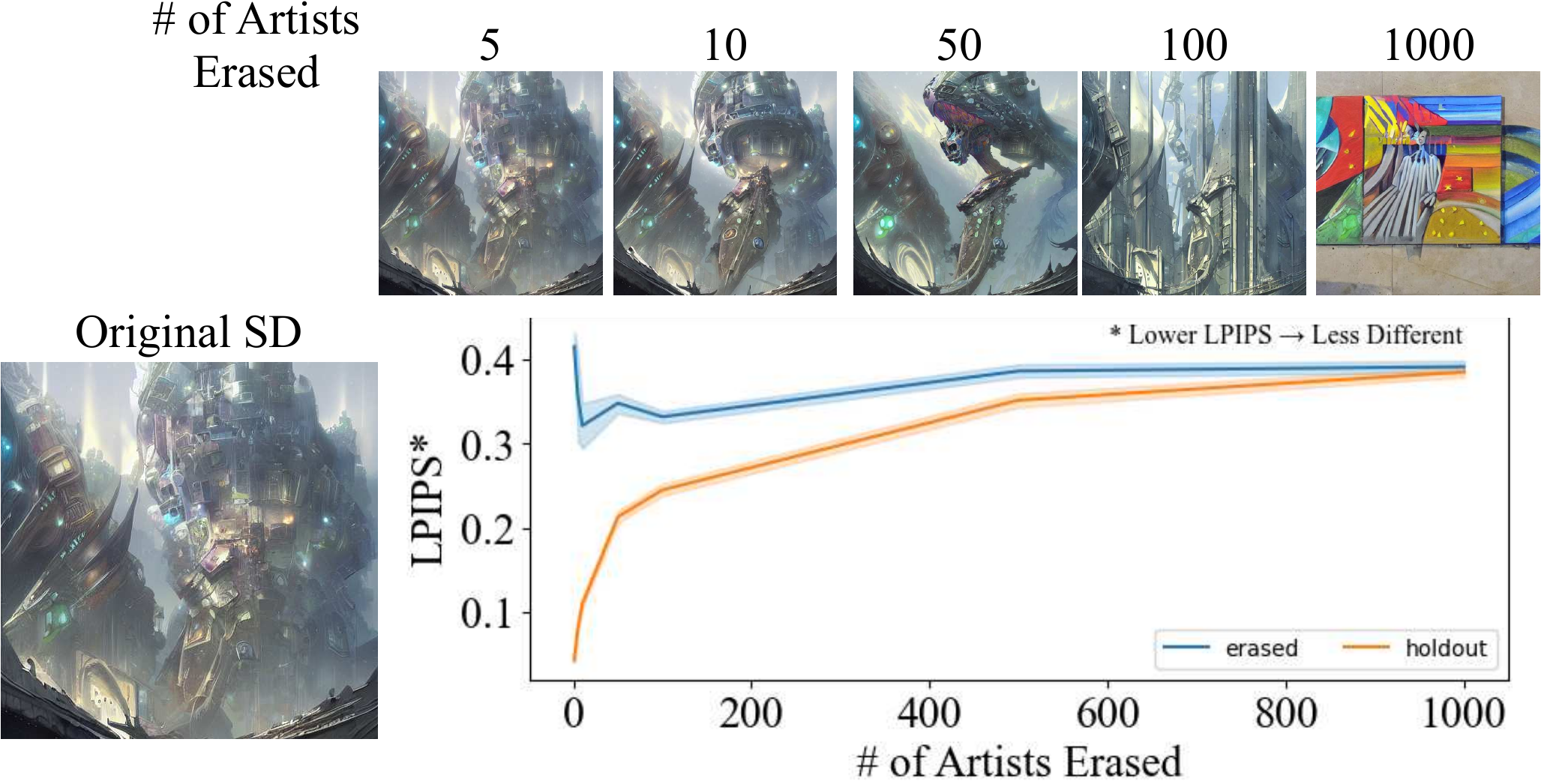}
    \caption{The samples demonstrate model performance on erased artists after editing. We observed that erasing over 100 artists begins negatively impacting output for holdout artists, as evidenced by the increasing LPIPS after 100 erasures. Erasing 50 or fewer artists resulted in negligible interference on non-erased concepts.}
    \label{fig:erase_limit}
\end{figure}

\begin{figure*}
    \centering
    \includegraphics[width=\linewidth]{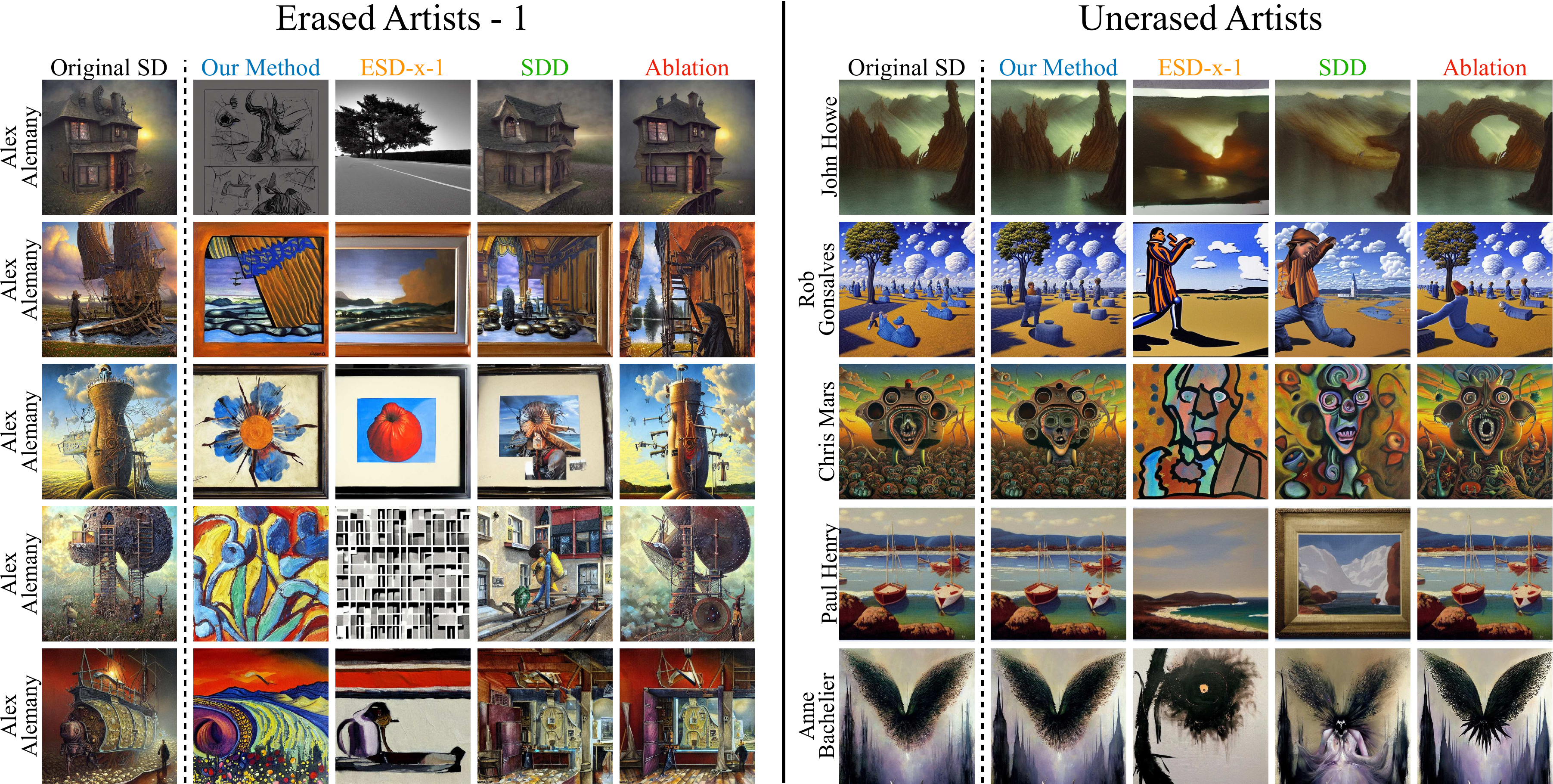}
    \caption{Our method demonstrates a complete erasure of the intended artistic style and the least interference with the holdout artists that were neither erased nor preserved.}
    \label{fig:style1}
\end{figure*}

\begin{figure*}
    \centering
    \includegraphics[width=\linewidth]{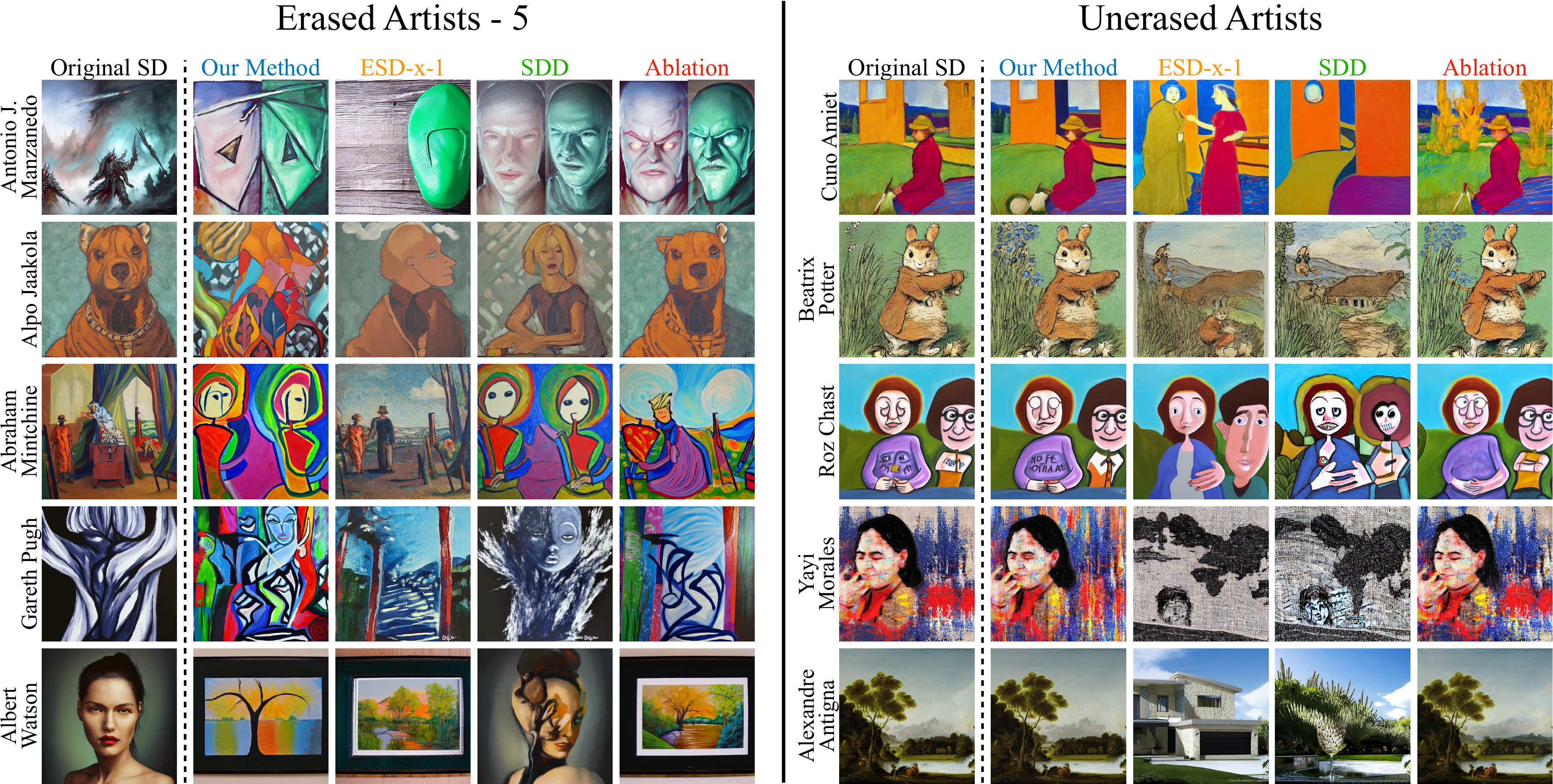}
    \caption{Our method demonstrates strong multi concept erasure of intended artistic styles and the least interference with the holdout artists that were neither erased nor preserved.}
    \label{fig:style2}
\end{figure*}

\begin{figure*}
    \centering
    \includegraphics[width=\linewidth]{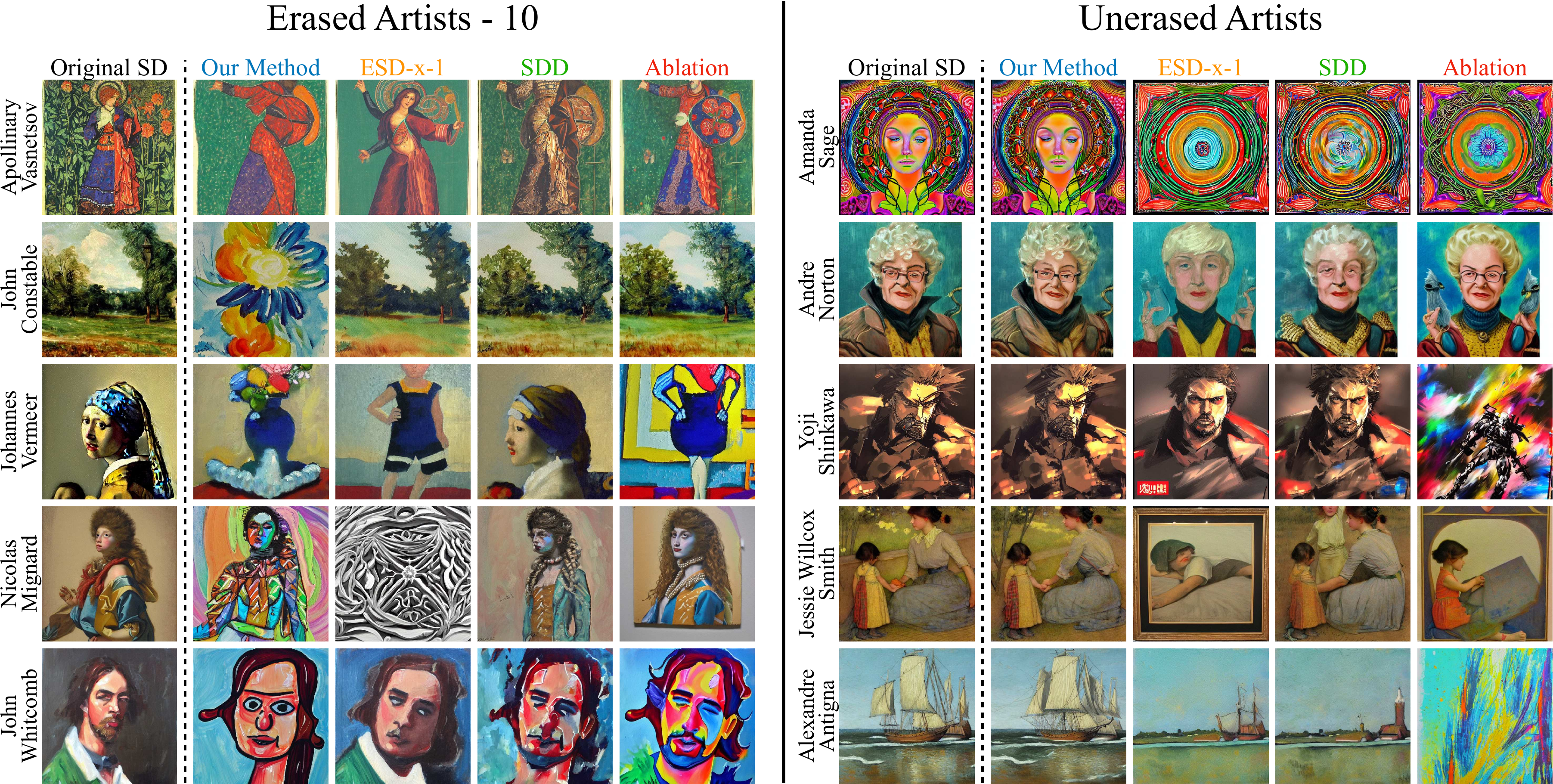}
    \caption{Our method demonstrates strong multi-concept erasure of intended artistic styles and the least interference with the holdout artists that were neither erased nor preserved. Previous methods start showing interference effects when erasing 10 artists}
    \label{fig:style3}
\end{figure*}

\begin{figure*}
    \centering
    \includegraphics[width=\linewidth]{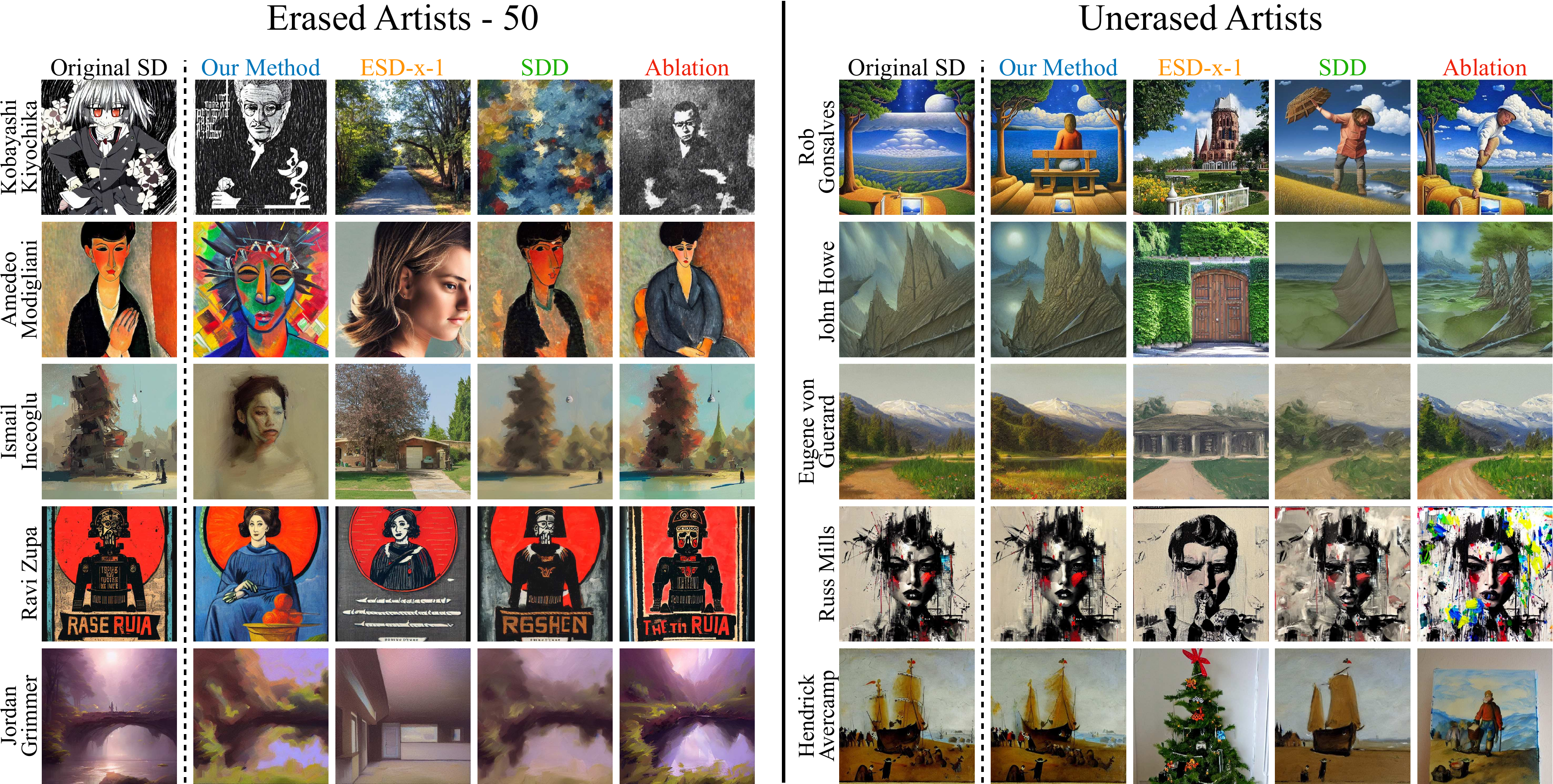}
    \caption{Our method demonstrates strong multi-concept erasure of intended artistic styles and the least interference with the holdout artists that were neither erased nor preserved. Previous methods start showing interference effects when erasing 50 artists}
    \label{fig:style4}
\end{figure*}

\begin{figure*}
    \centering
    \includegraphics[width=1\linewidth]{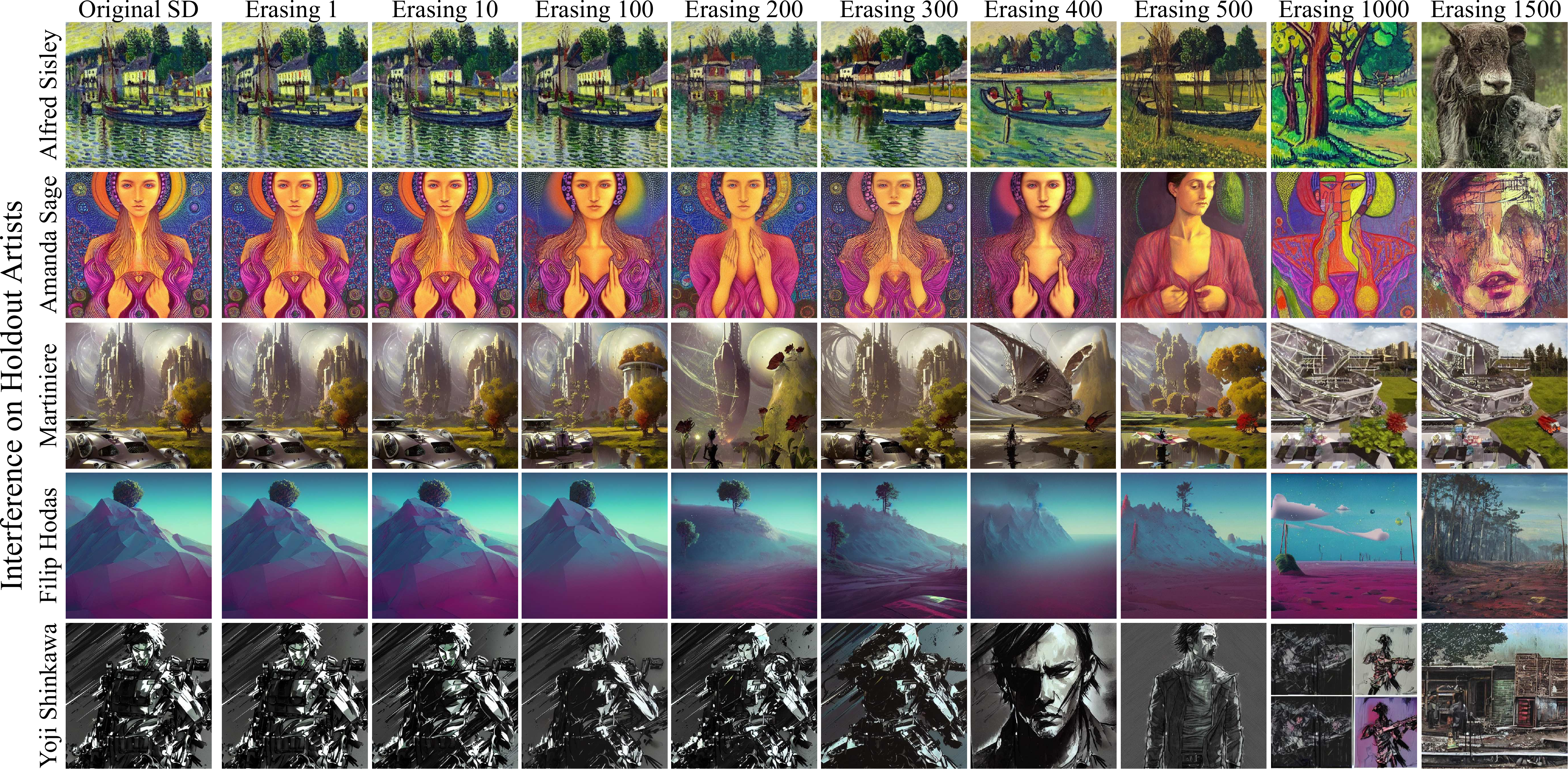}
    \caption{The samples demonstrate edited model performance on holdout artists. We observed changes in output quality for holdout styles after erasing 300 artists. At 1000 erasures, the network starts to lose the artistic nuance in its generated images.}
    \label{fig:limit_holdout}
\end{figure*}

We tested the limits of erasing artistic styles using our technique. As shown in Figure \ref{fig:erase_limit}, quality for holdout artists declines when erasing over 100 styles, evidenced by the increasing LPIPS after 100 erasures. With 50 or fewer erasures, interference was minimal for non-targeted concepts. Additional qualitative results for our method and baselines are provided in Figures \ref{fig:style1}-\ref{fig:style4}. The baselines are less effective at removing multiple artists and exhibit greater interference on unerased styles compared to our approach. Our method demonstrates superior erasure while minimizing interference when removing multiple artists. Figure~\ref{fig:limit_holdout} shows the results of stress testing the limits of artistic style erasure before general art capabilities decline. We observed the model starts to lose artistic nuance in generated outputs after approximately 1000 edits. \par
\begin{figure}
    \centering
    \includegraphics[width=1\linewidth]{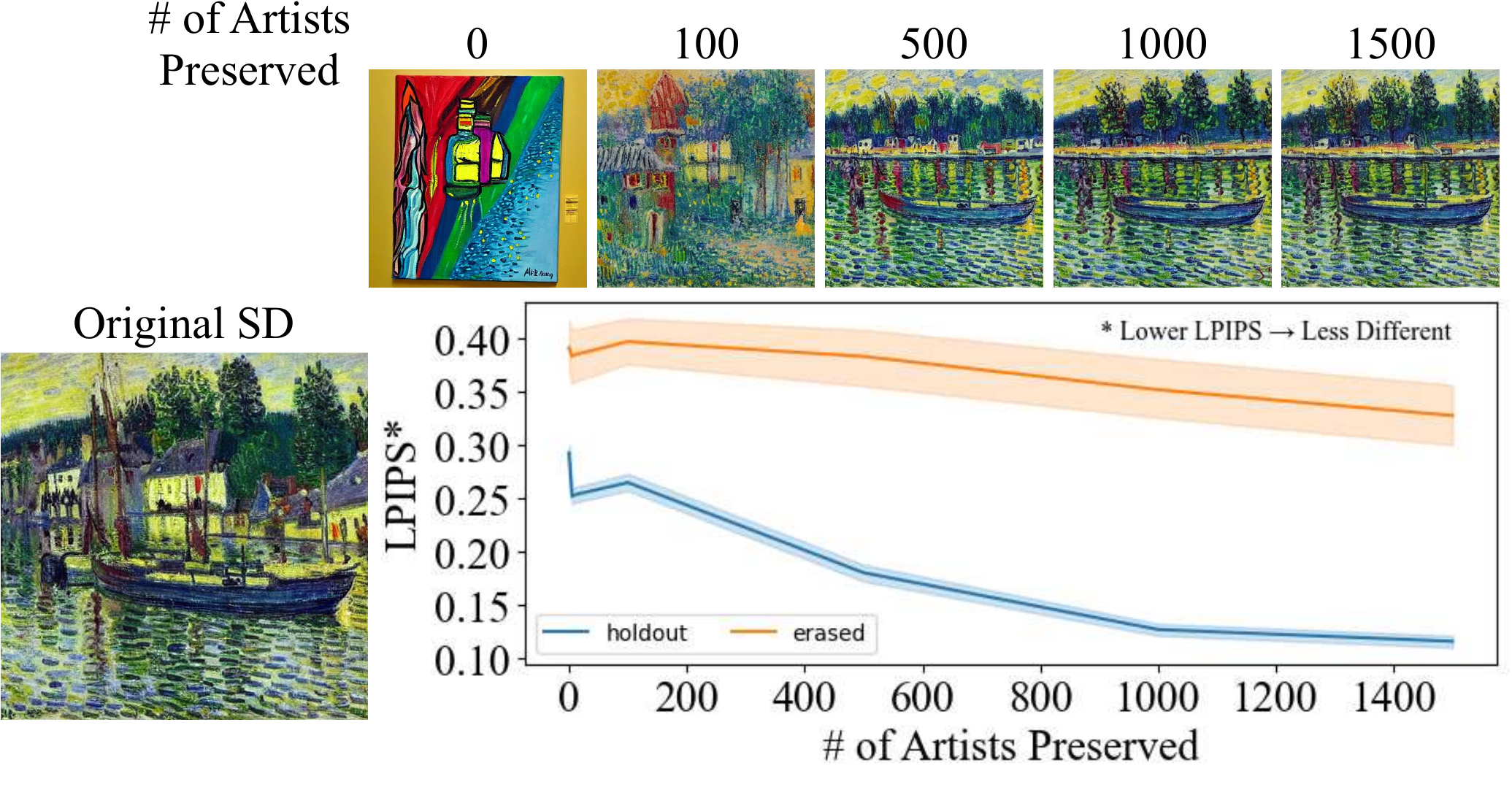}
    \caption{The samples show the performance of edited models on holdout artist. We observe that preserving more artists is beneficial for reducing model's interference with nearby concepts. The plot shows LPIPS between original SD and models erasing 10 random artists with variable number of preservation artists. We find that preserving 500 artists and more has close to no interference on other surrounding concepts when erasing 10 artists.}
    \label{fig:preserve_limit}
\end{figure}

An important follow-up question is the minimum number of artists requiring preservation to maintain performance when erasing styles. We analyzed this by testing preservation limits when erasing 10 artists, as shown in Figure \ref{fig:preserve_limit}. Erasing up to 1500 artists while preserving subsets, we assessed the impact on 100 non-preserved, non-erased artists. The LPIPS divergence indicates preserving at least 500 artists is essential for retaining model performance. Additional results and analysis are provided in the Appendix.

\begin{figure*}
    \centering
    \includegraphics[width=1\linewidth]{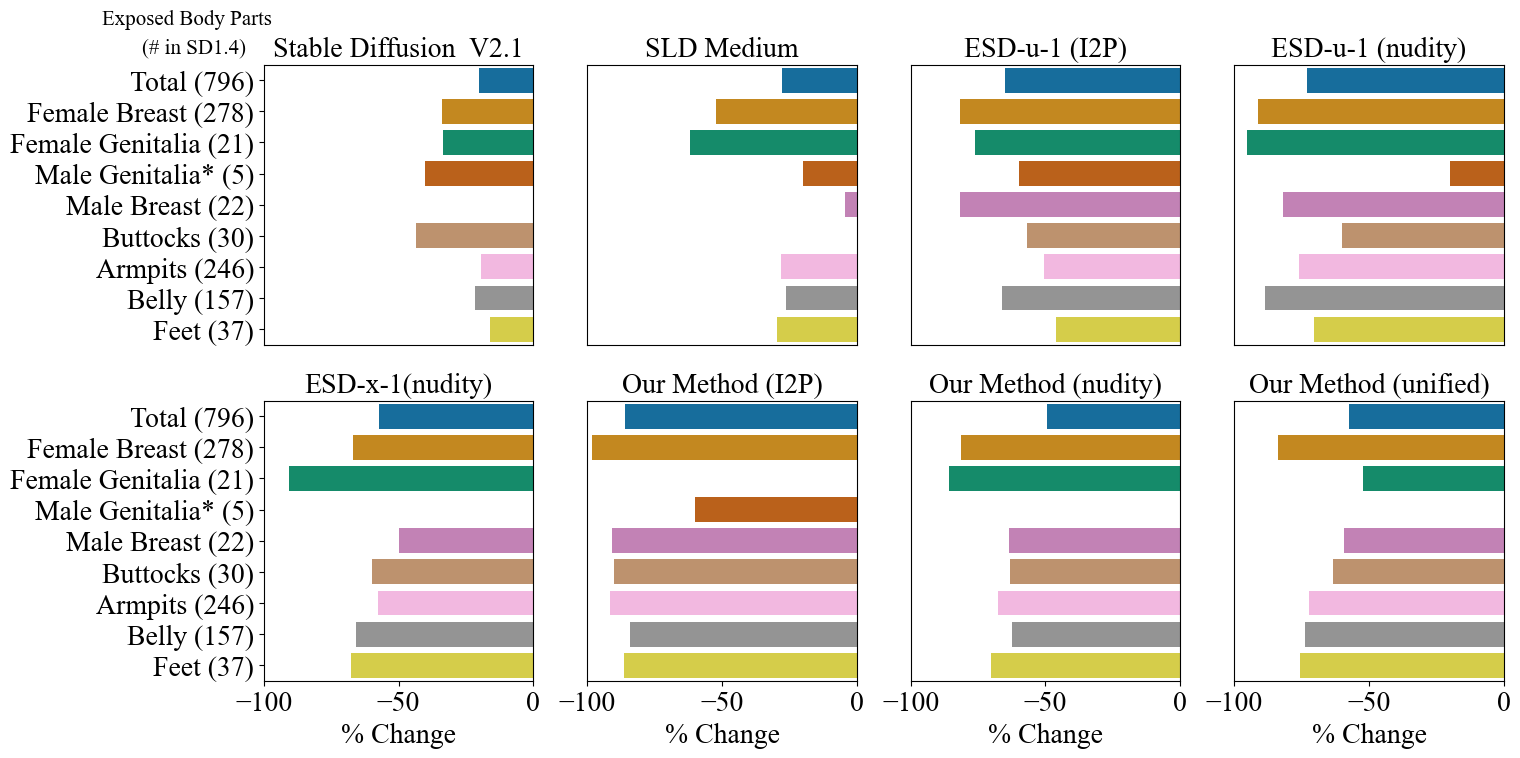}
    \caption{Our method erases nudity content from pre-trained SD and has an advantage of erasing multiple concepts in I2P prompts. The figure shows percentage reduction in nudity classified samples for each body part type on I2P prompts compared to SD. \texttt{"Nudity"} erased model performs very similar to \texttt{ESD-x-1} as both the methods edit only cross attentions. Although, as noted in main paper, we find that our method results in a more finer edit and has better alignment with COCO.}
    \label{fig:nudity_bar_full}
\end{figure*}

\begin{table}
    \centering
    \begin{tabular}{lcccc}
        \textbf{Profession} & \textbf{SD}  & \textbf{TIME +}&  \textbf{Ours} & \textbf{Ours} \tabularnewline
        
          &  & \textbf{Preserve} &  \textbf{Debiased} & \textbf{Unified} \tabularnewline

         \hline
         Assistant & 0.19 & 0.57  & 0.14 &  \textbf{0.09} \tabularnewline
         Cook & 0.82  & 0.15 & \textbf{0.03} &  0.14 \tabularnewline
         Worker & 1.00  & 0.15 & \textbf{0.06} &   0.09 \tabularnewline
         Analyst & 0.58  & 0.13  & 0.20  &   \textbf{0.03} \tabularnewline
         Doctor & 0.78  & 0.41  & 0.20  &   \textbf{0.09} \tabularnewline
         \cdashline{1-5}
         WinoBias & 0.67 & 0.31 & \textbf{0.22 } &  0.27  
    \end{tabular}
    \caption{Quantitative evaluation of profession debiasing for the unified model compared to an individual debiasing model. The metric $\Delta$ measures percentage deviation from equal gender ratios ($\Delta$=0 denotes perfect equality). On average, the unified model achieves comparable debiasing performance to the individually finetuned model.}
    \label{tab:gender_unified}
\end{table}

\subsection{Debiasing}
Table~\ref{tab:gender_unified} shows the performance of the unified model compared to our individual debias models. On an average we find that unified models show a similar performance to debiased models. \par
\begin{figure*}
    \centering
    \includegraphics[width=\linewidth]{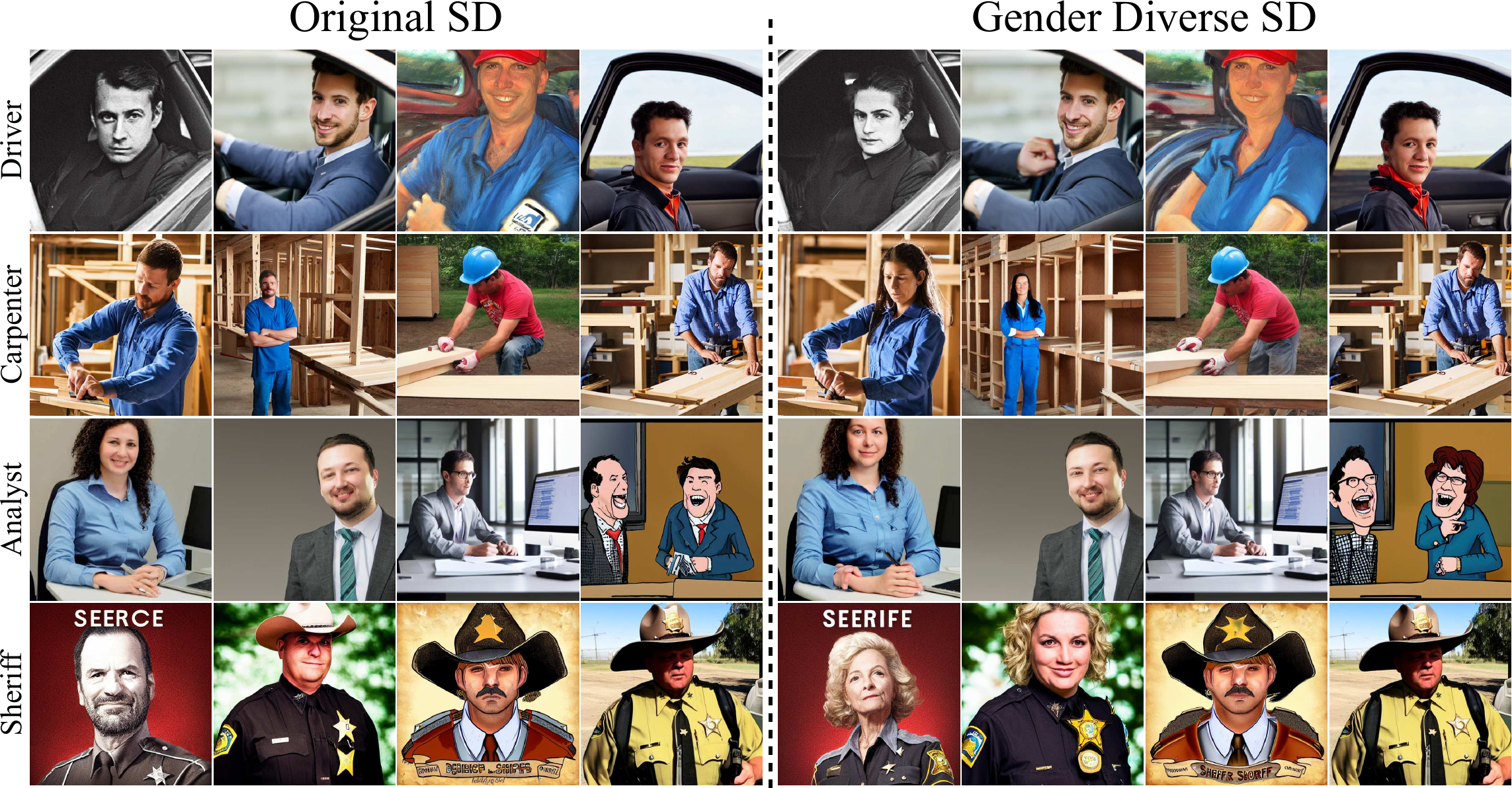}
    \caption{Our method improves the gender representation of professions in the stable diffusion generated images. We find that the images precisely change the gender while keeping the rest of the scene intact.}
    \label{fig:gender_appendix1}
\end{figure*}

\begin{figure*}
    \centering
    \includegraphics[width=\linewidth]{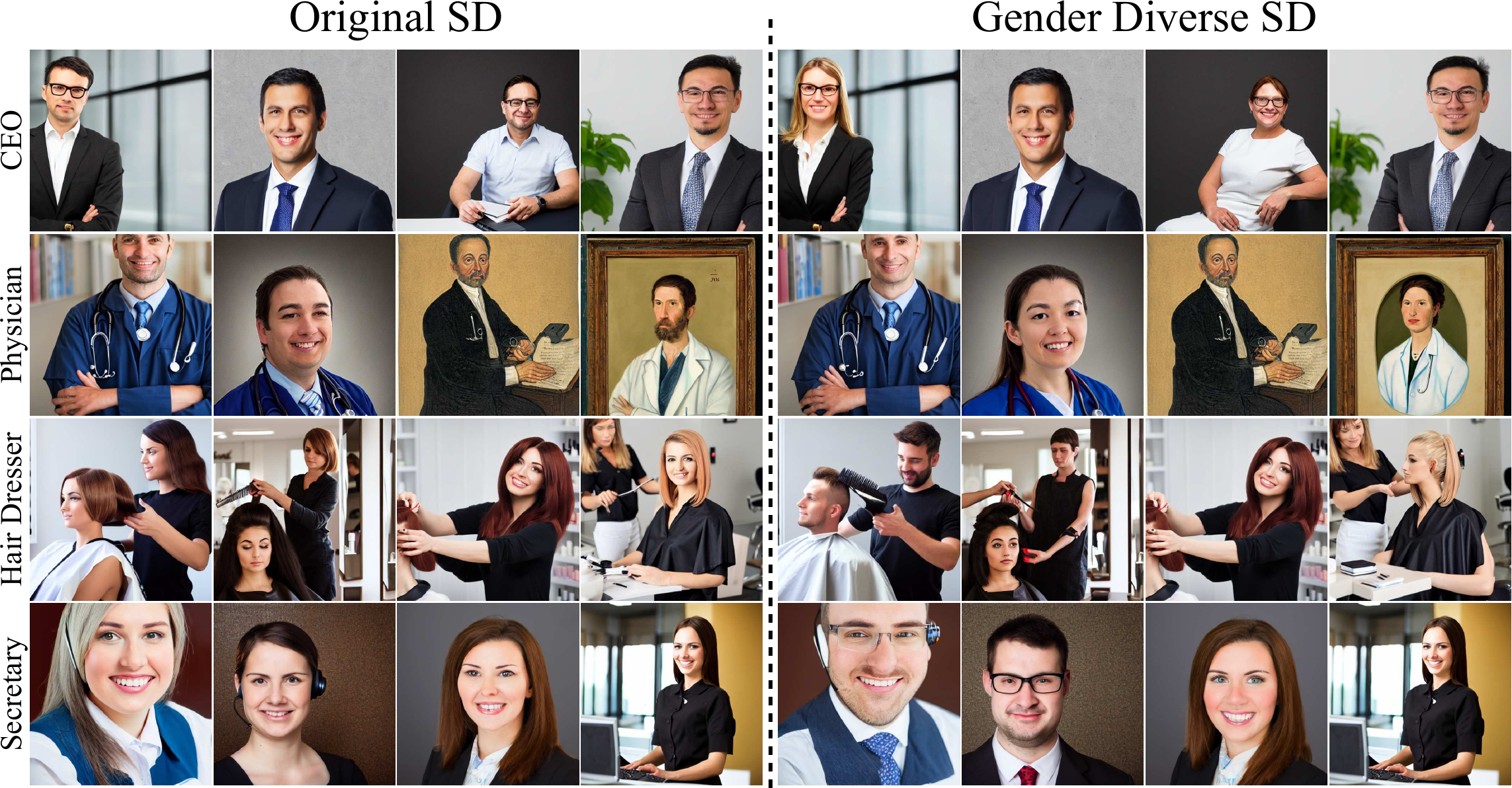}
    \caption{Our method improves the gender representation of professions in the stable diffusion generated images. We find that the images precisely change the gender while keeping the rest of the scene intact.}
    \label{fig:gender_appendix2}
\end{figure*}

\begin{figure*}
    \centering
    \includegraphics[width=\linewidth]{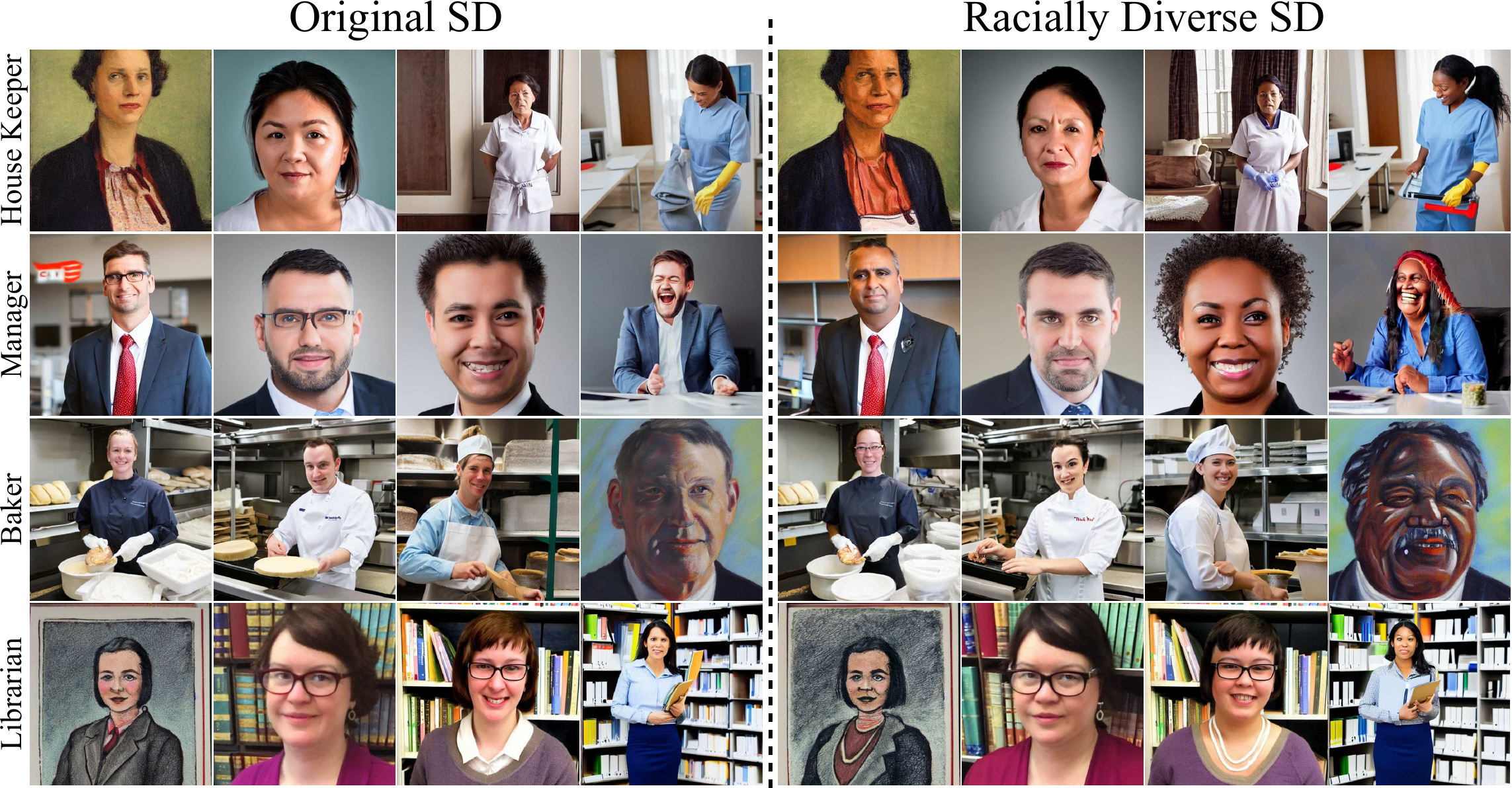}
    \caption{Our method improves the racial diversity of professions in the pre-trained stable diffusion. We show images from the original SD and the corresponding images from the edited model for the same prompts and seeds for comparison. We find that our edited model has a better race representation.}
    \label{fig:race_appendix1}
\end{figure*}

 Table \ref{tab:gender_bias_full} displays the debiased results for all 36 individual professions from the WinoBias dataset. Our method consistently reduced bias and increased gender diversity in Stable Diffusion outputs for most professions. Additional qualitative results demonstrating gender debiasing can be seen in Figures \ref{fig:gender_appendix1} and \ref{fig:gender_appendix2}. Figure \ref{fig:race_appendix1} provides further examples of improved racial diversity using our technique. \par 

The algorithm outlined in the main paper describes our approach for debiasing diffusion model concepts by iteratively editing cross-attention weights. It takes as input the concepts to edit $c_i$, concepts to preserve $c_j$, and attribute text prompts to debias $a_p$. In a loop, current attribute ratio distributions $R_{curr}$ are calculated for each concept using validation prompts and CLIP classification. $R_{curr}$ is an $m \times p$ matrix, where $m$ is the current edit list size and $p$ the attribute count. The debiasing constants $\alpha_p$ are then computed proportionally to the difference between current and desired ratios, scaled by learning rate $\eta$. Once a concept is sufficiently debiased (within 5\% of target), it is removed from the edit list and added to the preservation list. \par
This is done for 3 reasons - first, different concepts require varying levels of editing to remove bias, so they do not all debias at the same time. Once a concept is sufficiently debiased, we remove it from the edit list to avoid unnecessary generation of validation images for that concept. Second, editing one concept creates interference that can disrupt other concepts. Adding the debiased concept to the preservation list protects it from being affected by future edits. Third, careful asymmetric calculation of the debiasing constants $\alpha_p$ is required, unlike the symmetric constants used for erasing and moderating concepts. The optimal $\alpha_p$ values differ across concepts and attributes, necessitating the iterative tuning process.

\begin{table*}
    \centering
    \begin{tabular}{lcccccc}
        \textbf{Profession} & \textbf{Original-SD}  & \textbf{Concept Algebra} & \textbf{Debias-VL} & \textbf{TIME} & \textbf{TIME + Preserve} & \textbf{Ours} \tabularnewline
        
         \hline
        Attendant & 0.13 $\pm$ 0.06 & 0.23 $\pm$ 0.08 & 0.30 $\pm$ 0.04 & 0.50 $\pm$ 0.01 & 0.38 $\pm$ 0.11 & 0.09 $\pm$ 0.04 \tabularnewline
        Cashier & 0.67 $\pm$ 0.04 & 0.71 $\pm$ 0.10 & 0.23 $\pm$ 0.07 & 0.46 $\pm$ 0.01 & 0.23 $\pm$ 0.15 & 0.16 $\pm$ 0.06 \tabularnewline
        Teacher & 0.42 $\pm$ 0.01 & 0.46 $\pm$ 0.00 & 0.11 $\pm$ 0.05 & 0.34 $\pm$ 0.06 & 0.07 $\pm$ 0.06 & 0.06 $\pm$ 0.02 \tabularnewline
        Nurse & 0.99 $\pm$ 0.01 & 0.91 $\pm$ 0.05 & 0.87 $\pm$ 0.01 & 0.34 $\pm$ 0.03 & 0.30 $\pm$ 0.07 & 0.39 $\pm$ 0.07 \tabularnewline
        Assistant & 0.19 $\pm$ 0.05 & 0.20 $\pm$ 0.07 & 0.35 $\pm$ 0.15 & 0.32 $\pm$ 0.06 & 0.57 $\pm$ 0.08 & 0.14 $\pm$ 0.06 \tabularnewline
        Secretary & 0.88 $\pm$ 0.01 & 0.65 $\pm$ 0.07 & 0.65 $\pm$ 0.01 & 0.58 $\pm$ 0.09 & 0.71 $\pm$ 0.02 & 0.10 $\pm$ 0.10 \tabularnewline
        Cleaner & 0.38 $\pm$ 0.04 & 0.11 $\pm$ 0.06 & 0.18 $\pm$ 0.04 & 0.58 $\pm$ 0.07 & 0.79 $\pm$ 0.04 & 0.33 $\pm$ 0.07 \tabularnewline
        Receptionist & 0.99 $\pm$ 0.01 & 0.90 $\pm$ 0.08 & 0.74 $\pm$ 0.04 & 0.36 $\pm$ 0.10 & 0.24 $\pm$ 0.12 & 0.38 $\pm$ 0.01 \tabularnewline
        Clerk & 0.10 $\pm$ 0.07 & 0.11 $\pm$ 0.08 & 0.10 $\pm$ 0.04 & 0.40 $\pm$ 0.03 & 0.76 $\pm$ 0.05 & 0.23 $\pm$ 0.06 \tabularnewline
        Counselor & 0.06 $\pm$ 0.05 & 0.30 $\pm$ 0.03 & 0.10 $\pm$ 0.07 & 0.74 $\pm$ 0.08 & 0.41 $\pm$ 0.06 & 0.40 $\pm$ 0.02 \tabularnewline
        Designer & 0.23 $\pm$ 0.05 & 0.25 $\pm$ 0.12 & 0.48 $\pm$ 0.06 & 0.44 $\pm$ 0.06 & 0.23 $\pm$ 0.16 & 0.07 $\pm$ 0.05 \tabularnewline
        Hairdresser & 0.74 $\pm$ 0.11 & 0.37 $\pm$ 0.16 & 0.61 $\pm$ 0.04 & 0.32 $\pm$ 0.01 & 0.41 $\pm$ 0.09 & 0.16 $\pm$ 0.04 \tabularnewline
        Writer & 0.15 $\pm$ 0.03 & 0.07 $\pm$ 0.03 & 0.45 $\pm$ 0.04 & 0.54 $\pm$ 0.08 & 0.52 $\pm$ 0.08 & 0.31 $\pm$ 0.08 \tabularnewline
        Housekeeper & 0.93 $\pm$ 0.04 & 0.68 $\pm$ 0.18 & 0.80 $\pm$ 0.07 & 0.32 $\pm$ 0.03 & 0.68 $\pm$ 0.07 & 0.41 $\pm$ 0.05 \tabularnewline
        Baker & 0.81 $\pm$ 0.01 & 0.19 $\pm$ 0.04 & 0.72 $\pm$ 0.05 & 0.40 $\pm$ 0.04 & 0.19 $\pm$ 0.14 & 0.29 $\pm$ 0.08 \tabularnewline
        Librarian & 0.86 $\pm$ 0.06 & 0.66 $\pm$ 0.07 & 0.34 $\pm$ 0.06 & 0.26 $\pm$ 0.05 & 0.35 $\pm$ 0.01 & 0.07 $\pm$ 0.07 \tabularnewline
        Tailor & 0.30 $\pm$ 0.01 & 0.21 $\pm$ 0.05 & 0.33 $\pm$ 0.11 & 0.50 $\pm$ 0.03 & 0.03 $\pm$ 0.00 & 0.27 $\pm$ 0.01 \tabularnewline
        Driver & 0.97 $\pm$ 0.02 & 0.20 $\pm$ 0.07 & 0.65 $\pm$ 0.04 & 0.48 $\pm$ 0.09 & 0.17 $\pm$ 0.08 & 0.21 $\pm$ 0.07 \tabularnewline
        Supervisor & 0.50 $\pm$ 0.01 & 0.07 $\pm$ 0.03 & 0.43 $\pm$ 0.04 & 0.50 $\pm$ 0.07 & 0.42 $\pm$ 0.03 & 0.26 $\pm$ 0.04 \tabularnewline
        Janitor & 0.91 $\pm$ 0.05 & 0.71 $\pm$ 0.06 & 0.75 $\pm$ 0.05 & 0.36 $\pm$ 0.08 & 0.47 $\pm$ 0.12 & 0.16 $\pm$ 0.04 \tabularnewline
        Cook & 0.82 $\pm$ 0.04 & 0.48 $\pm$ 0.16 & 0.52 $\pm$ 0.07 & 0.38 $\pm$ 0.03 & 0.15 $\pm$ 0.10 & 0.03 $\pm$ 0.02 \tabularnewline
        Laborer & 0.99 $\pm$ 0.01 & 0.81 $\pm$ 0.06 & 0.98 $\pm$ 0.03 & 0.48 $\pm$ 0.08 & 0.24 $\pm$ 0.09 & 0.09 $\pm$ 0.02 \tabularnewline
        Constr. worker & 1.00 $\pm$ 0.00 & 0.95 $\pm$ 0.01 & 1.00 $\pm$ 0.00 & 0.40 $\pm$ 0.01 & 0.15 $\pm$ 0.05 & 0.06 $\pm$ 0.04 \tabularnewline
        Developer & 0.90 $\pm$ 0.03 & 0.74 $\pm$ 0.02 & 0.90 $\pm$ 0.04 & 0.50 $\pm$ 0.01 & 0.47 $\pm$ 0.07 & 0.51 $\pm$ 0.02 \tabularnewline
        Carpenter & 0.92 $\pm$ 0.05 & 0.84 $\pm$ 0.01 & 0.98 $\pm$ 0.01 & 0.52 $\pm$ 0.06 & 0.52 $\pm$ 0.05 & 0.06 $\pm$ 0.02 \tabularnewline
        Manager & 0.54 $\pm$ 0.06 & 0.15 $\pm$ 0.01 & 0.30 $\pm$ 0.05 & 0.38 $\pm$ 0.05 & 0.15 $\pm$ 0.01 & 0.19 $\pm$ 0.07 \tabularnewline
        Lawyer & 0.46 $\pm$ 0.08 & 0.13 $\pm$ 0.06 & 0.52 $\pm$ 0.05 &  0.64 $\pm$ 0.03 & 0.15 $\pm$ 0.03 & 0.30 $\pm$ 0.07 \tabularnewline
        Farmer & 0.97 $\pm$ 0.02 & 0.58 $\pm$ 0.09 & 0.97 $\pm$ 0.02 & 0.46 $\pm$ 0.02 & 0.27 $\pm$ 0.08 & 0.41 $\pm$ 0.01 \tabularnewline
        Salesperson & 0.60 $\pm$ 0.08 & 0.18 $\pm$ 0.05 & 0.07 $\pm$ 0.05 & 0.52 $\pm$ 0.05 & 0.05 $\pm$ 0.01 & 0.38 $\pm$ 0.05 \tabularnewline
        Physician & 0.62 $\pm$ 0.14 & 0.36 $\pm$ 0.10 & 0.70 $\pm$ 0.07 & 0.56 $\pm$ 0.06 & 0.49 $\pm$ 0.04 & 0.42 $\pm$ 0.01 \tabularnewline
        Guard & 0.86 $\pm$ 0.02 & 0.43 $\pm$ 0.12 & 0.48 $\pm$ 0.06 & 0.30 $\pm$ 0.10 & 0.10 $\pm$ 0.12 & 0.12 $\pm$ 0.07 \tabularnewline
        Analyst & 0.58 $\pm$ 0.12 & 0.24 $\pm$ 0.18 & 0.71 $\pm$ 0.02 & 0.52 $\pm$ 0.03 & 0.13 $\pm$ 0.05 & 0.20 $\pm$ 0.07 \tabularnewline
        Mechanic & 0.99 $\pm$ 0.01 & 0.65 $\pm$ 0.04 & 0.92 $\pm$ 0.01 & 0.38 $\pm$ 0.09 & 0.21 $\pm$ 0.04 & 0.23 $\pm$ 0.08 \tabularnewline
        Sheriff & 0.99 $\pm$ 0.01 & 0.38 $\pm$ 0.22 & 0.82 $\pm$ 0.08 & 0.22 $\pm$ 0.05 & 0.10 $\pm$ 0.05 & 0.10 $\pm$ 0.03 \tabularnewline
        Ceo & 0.87 $\pm$ 0.03 & 0.25 $\pm$ 0.11 & 0.37 $\pm$ 0.11 & 0.28 $\pm$ 0.04 & 0.18 $\pm$ 0.05 & 0.28 $\pm$ 0.03 \tabularnewline
        Doctor & 0.78 $\pm$ 0.04 & 0.40 $\pm$ 0.02 & 0.50 $\pm$ 0.04 & 0.58 $\pm$ 0.03 & 0.41 $\pm$ 0.08 & 0.20 $\pm$ 0.02 \tabularnewline
         \cdashline{1-7}
         WinoBias & 0.67 $\pm$ 0.01 & 0.43 $\pm$ 0.01 & 0.55 $\pm$ 0.01 & 0.44 $\pm$ 0.00 & 0.31 $\pm$ 0.00 & 0.22 $\pm$ 0.00
    \end{tabular}
    \caption{Our method has a consistent debiasing performance compared to previous inference and model editing methods. The presented metric $\Delta$ measures the percentage deviation from equal ratios ($\Delta=0$ indicates perfect equal distribution across attributes) on 5 randomly picked professions out of 36 from the WinoBias dataset. On average, our method has the least deviation from the desired distribution.}
    \label{tab:gender_bias_full}
\end{table*}

\subsection{Moderating NSFW}
Figure \ref{fig:nudity_bar_full} displays the detailed erasure effects on different nudity classes classified by Nudenet. Our method demonstrates similar erasure to ESD-x for individual classes, while showing less interference on other concepts. The major advantage of our technique emerges in multi-concept erasure for I2P prompts and overall NSFW moderation, where our approach erases better than ESD methods across different NSFW classes.

\subsection{Erasing Objects}
\begin{figure}
    \centering
    \includegraphics[width=\linewidth]{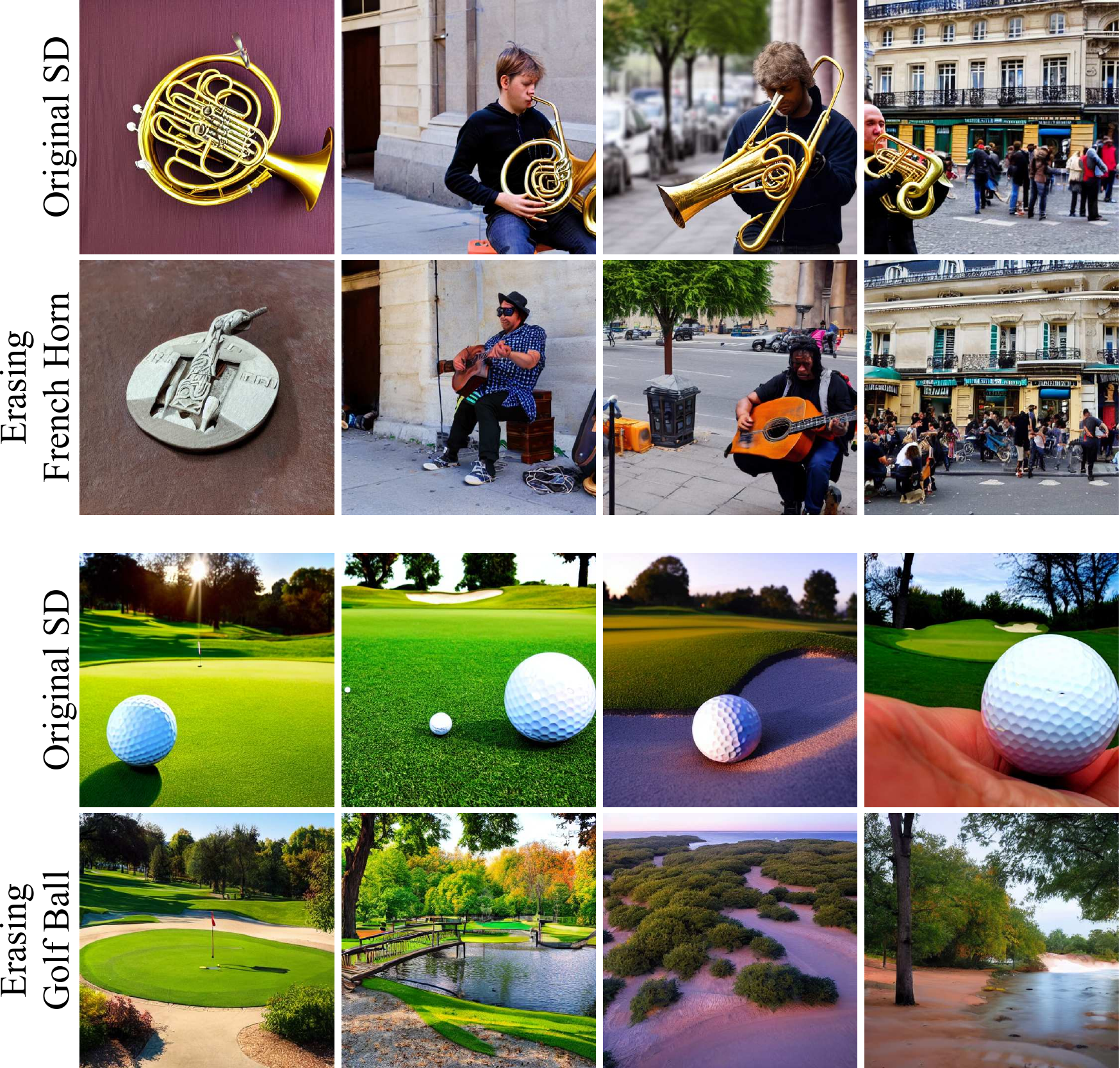}
    \caption{Our method demonstrates a complete erasure of the intended object and the least interference with unerased objects that are not explicitly preserved.}
    \label{fig:object1}
\end{figure}

\begin{figure}
    \centering
    \includegraphics[width=\linewidth]{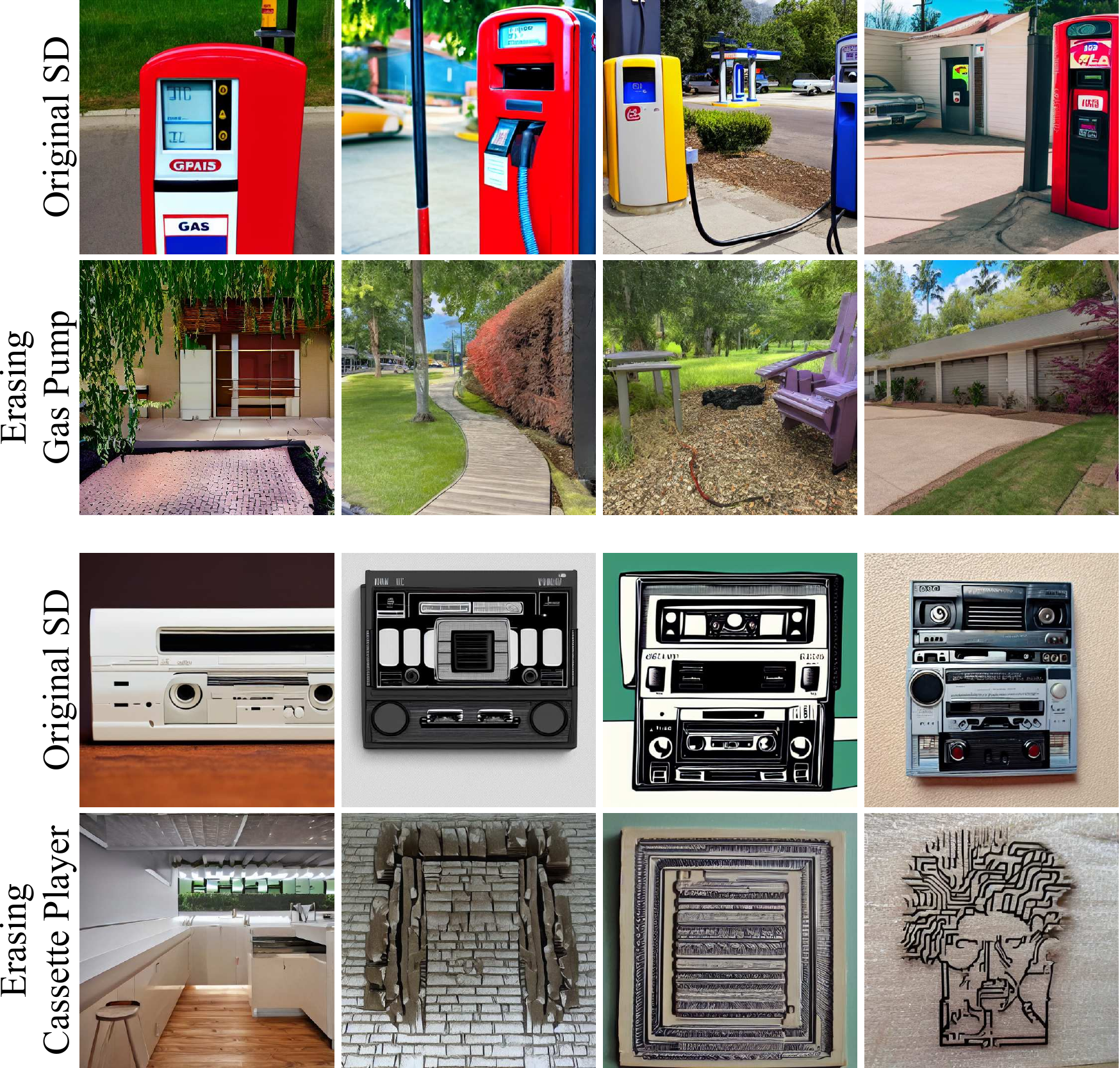}
    \caption{Our method demonstrates a complete erasure of the intended object and the least interference with unerased objects that are not explicitly preserved.}
    \label{fig:object2}
\end{figure}

\begin{figure}
    \centering
    \includegraphics[width=\linewidth]{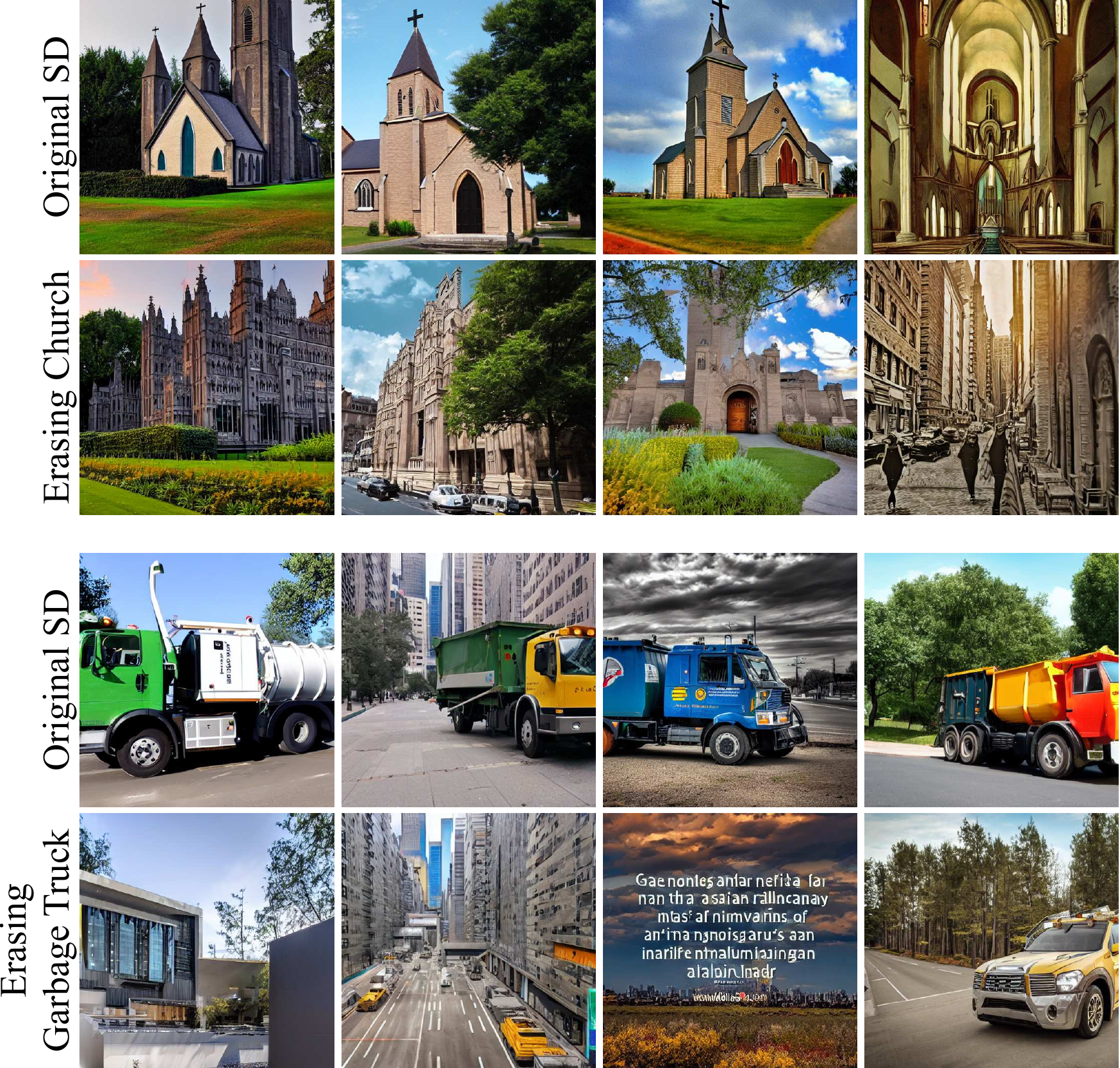}
    \caption{Our method demonstrates a complete erasure of the intended object and the least interference with unerased objects that are not explicitly preserved.}
    \label{fig:object3}
\end{figure}

Figures~\ref{fig:object1}-~\ref{fig:object3} demonstrate effective object erasure using our method. One limitation of ESD-u was only partial removal of objects like churches, where major attributes such as crosses and tinted windows were erased but the building remained. In contrast, our approach shows stronger editing that clearly erases the full object. 

\end{document}